\documentclass[11pt]{article}
\usepackage[english]{babel}
\usepackage{amssymb,amsmath,graphicx}
\usepackage{amsthm}
\usepackage{float}
\usepackage{afterpage}
\usepackage{dcolumn}
\usepackage{multirow}
\usepackage{color}
\usepackage[round,authoryear]{natbib}
\usepackage{enumitem}
\usepackage{slashbox}
\usepackage{pict2e}
\usepackage{setspace}
\usepackage{url}

\usepackage[norelsize,boxed,noend]{algorithm2e}
\SetKw{Continue}{continue}
\SetAlCapSkip{0.5em}
\DontPrintSemicolon
\SetKwInput{KwData}{Input}
\SetKwInput{KwResult}{Output}

\usepackage{rotating}
\usepackage[table]{xcolor}
\usepackage{float}
\usepackage{color} 
\usepackage{multirow}
\usepackage{graphicx}
\usepackage{flafter}
\usepackage{natbib}
\usepackage{microtype}
\usepackage[table]{xcolor}
\usepackage{tikz}
\usepackage[justification=centering]{caption}
\usepackage[all]{nowidow}

\DeclareGraphicsExtensions{.png}
\newcolumntype{M}[1]{>{\centering}m{#1}}
\newcolumntype{H}{>{\setbox0=\hbox\bgroup}c<{\egroup}@{}}

\renewcommand{\arraystretch}{1.1}

 \bibpunct[, ]{(}{)}{,}{a}{}{,}%
 \def\bibsep{\smallskipamount}%

\newcommand{\myFont}[1]{{\ttfamily\fontseries{b}\selectfont#1}}

\newtheorem{theorem}{Theorem}

\newcolumntype{d}[1]{D{.}{.}{#1}}

\newcommand{\Nat}{{\mathbb{N}}}
\newcommand{\cG}{{\mathcal G}}
\newcommand{\cV}{{\mathcal V}}

\newcommand{\cE}{{\mathcal E}}

\newcommand{\cQ}{{\mathcal Q}}

\newcommand{\myblue}[1]{#1}

\title{Large Neighborhood-Based Metaheuristic and Branch-and-Price for the Pickup and Delivery Problem with Split Loads}

\author{Matheus Nohra Haddad, Luiz Satoru Ochi, Marcone Jamilson Freitas Souza, Richard Hartl, Simone Martins, Thibaut Vidal}

\begin{document}

\begin{center}

\begin{LARGE}
Large Neighborhood-Based Metaheuristic and Branch-and-Price   \vspace*{0.2cm}

for the Pickup and Delivery Problem with Split Loads
\end{LARGE}

\vspace*{0.6cm}

\textbf{Matheus Nohra Haddad} \\
Instituto de Computa\c{c}\~ao -- Universidade Federal Fluminense \\
Rua Passo da P\'{a}tria, 156 - S\~{a}o Domingos, Niter\'{o}i - RJ, 24210-240, Brazil \\
mathaddad@gmail.com \\
\vspace*{0.35cm}
\textbf{Rafael Martinelli} \\
Departamento de Engenharia Industrial -- Pontif\'{i}cia Universidade Cat\'{o}lica do Rio de Janeiro \\
Rua Marqu\^{e}s de S\~{a}o Vicente, 225 - G\'{a}vea, Rio de Janeiro - RJ, 22451-900, Brazil \\
martinelli@puc-rio.br \\
\vspace*{0.35cm}
\textbf{Thibaut Vidal} \\
Departamento de Inform\'{a}tica -- Pontif\'{i}cia Universidade Cat\'{o}lica do Rio de Janeiro \\
Rua Marqu\^{e}s de S\~{a}o Vicente, 225 - G\'{a}vea, Rio de Janeiro - RJ, 22451-900, Brazil \\
vidalt@inf.puc-rio.br \\
\vspace*{0.35cm}
\textbf{Simone Martins, Luiz Satoru Ochi} \\
Instituto de Computa\c{c}\~ao -- Universidade Federal Fluminense \\
Rua Passo da P\'{a}tria, 156 - S\~{a}o Domingos, Niter\'{o}i - RJ, 24210-240, Brazil \\
simone@ic.uff.br, luiz.satoru@gmail.com \\
\vspace*{0.35cm}
\textbf{Marcone Jamilson Freitas Souza} \\
Departamento de Computa\c{c}\~{a}o -- Universidade Federal de Ouro Preto \\
Campus Universit\'{a}rio, Morro do Cruzeiro, Ouro Preto - MG, 35400-000, Brazil \\
marcone.freitas@gmail.com \\
\vspace*{0.35cm}
\textbf{Richard Hartl} \\
Department of Business Administration -- Universit\"at Wien \\
Oskar-Morgenstern-Platz 1, A-1090, Vienna, Austria \\
richard.hartl@univie.ac.at \\

\vspace*{0.3cm}

\begin{large}
Technical Report -- Universidade Federal Fluminense -- Feb 2018
\end{large}

\vspace*{0.2cm}

\end{center}

\noindent
\textbf{Abstract.} We consider the multi-vehicle one-to-one pickup and delivery problem with split loads, a NP-hard problem linked with a variety of applications for bulk product transportation, bike-sharing systems and inventory re-balancing. This problem is notoriously difficult due to the interaction of two challenging vehicle routing attributes, ``pickups and deliveries'' and ``split deliveries''. \myblue{This possibly leads to optimal solutions of a size that grows exponentially with the instance size, containing multiple visits per customer pair, even in the same route}. To solve this problem, we propose an iterated local search metaheuristic \myblue{as well as a branch-and-price algorithm}. The core of the metaheuristic consists of a new large neighborhood search, which reduces the problem of finding the best insertion combination of a pickup and delivery pair into a route (with possible splits) to a resource-constrained shortest path and knapsack problem. \myblue{Similarly, the branch-and-price 
algorithm uses sophisticated labeling techniques, route relaxations, pre-processing and branching rules for an efficient resolution.} Our computational experiments on classical single-vehicle instances demonstrate the excellent performance of the metaheuristic, which produces new best known solutions for 92 out of 93 test instances, and outperforms all previous algorithms. Experimental results on new multi-vehicle instances with distance constraints are also reported.  \myblue{The branch-and-price algorithm produces optimal solutions for instances with up to 20 pickup-and-delivery pairs, \myblue{and very accurate solutions are found by the metaheuristic}.}
\vspace*{0.2cm}

\vspace*{0.2cm}

\section{Introduction}
\label{sec:intro}

The classical vehicle routing problem (VRP) aims to find minimum-distance itineraries to service a set of geographically distributed customers with a fleet of vehicles, in such a way that each customer is visited once and the capacity of each vehicle is respected. This important combinatorial optimization problem has been the focus of extensive research since the 1960's \citep{Laporte09,Vidal13,Laporte2014a}. Over the years,  the classical version of the problem has been increasingly-well solved, but as new applications are discovered, many additional constraints, objectives, and other decision subsets, called ``attributes'' in \cite{Vidal13}, are combined with the classical problem, leading to new challenges.

A classical restriction of the VRP is that each delivery is done in one block by a single vehicle. \cite{Dror89} raised this restriction, allowing the total demand of a customer to be served during several visits, leading to the split delivery vehicle routing problem (SDVRP). At first, one might think that allowing split deliveries results in increased costs since more visits may be performed. Yet, this relaxation leads to a larger set of solutions, possibly opening the way to lower costs. The SDVRP is known to be notoriously more difficult to solve than the classical VRP from an exact method standpoint, and requires more sophisticated classes of neighborhoods to be adequately solved via metaheuristics \citep{Silva15}.

In the meantime, another VRP variant has drawn considerable attention over the years: the one-to-one pickup and delivery (PDP) problem, which requires to perform each service as a pair, in such a way that each pickup precedes its associated delivery in the same route. We refer to \cite{Berbeglia07} and \cite{Parragh08} for a detailed survey on pickup and delivery problems. Once again, an essential ingredient of state-of-the-art heuristics for this problem is the efficient exploration of a variety of neighbor solutions during the search, a task which tends to be more complex when pairs of deliveries are relocated or exchanged instead of single visits.

As both ``pickups and deliveries'' and ``split deliveries'' attributes require \myblue{sophisticated search techniques,} combining them into one vehicle routing variant poses significant methodological challenges, thus partly explaining the reduced number of methods proposed for the multi-vehicle one-to-one pickup and delivery problem with split loads (MPDPSL) despite its practical relevance. To this date, two main articles have considered this variant. \cite{Nowak08} presented a practical application faced by a logistic company which provides outsourced services. Then, \cite{Sahin13} solved this problem for the transportation of bulk products by ship. In this later case, each load is already packaged into multiple containers, and mail services collect and deliver multiple packets between origin and destination pairs. Finally, this problem is of high relevance in many other application contexts involving pickups and deliveries of divisible products, e.g., inventory between supermarkets \citep{Hartl2015}, or 
bike repositioning for self-sharing bike sharing systems \citep{Chemla2013a}.

Note that, despite the description of a multi-vehicle algorithm in \cite{Sahin13}, all previous experimental analyses have been restricted to single-vehicle cases. Indeed, the existing instances did not contain resource constraints on routes (e.g., distance or time), and the depot does not intervene as a replenishment facility in a one-to-one PDP. When the triangle inequality holds, it is never necessary or profitable to return to the depot, such that the search can be limited to single-vehicle solutions without compromising solution quality.

In this paper, we pursue the research on this difficult problem, by proposing new \myblue{heuristic and exact solution approaches}, along with experimental analyses on distance-constrained multi-vehicle benchmark instances. More precisely, we introduce a hybrid metaheuristic based on iterated local search (ILS) with random variable neighborhood descent (RVND), which incorporates classical neighborhoods and perturbation procedures with new larger dynamic programming-based neighborhoods for joint service reinsertions and optimization of split loads. \myblue{This method will be called  ILS--PDSL (ILS for pickup-and-delivery problems with split loads).}
\myblue{Moreover, we propose the first efficient branch-and-price algorithm for the MPDPSL. The method exploits problem-tailored route relaxations, pricing algorithms, pre-processing techniques and branching rules, allowing to solve problem instances with up to 20 pickup-and-delivery pairs.} As such, the key contributions of this work are:
\begin{enumerate}[nosep]
\item \myblue{Efficient heuristic and exact solutions approaches} for the MPDPSL;
\item new dynamic programming based neighborhoods for split pickup-and-delivery problems;
\item new state-of-the-art results for single-vehicle benchmark instances; and finally,
\item experimental analyses on new multi-vehicle benchmark instances.
\end{enumerate}

\section{Problem statement}\label{sec:problem}

Consider a graph $G = (V,E)$, where $V = P \cup D \cup \{0,2n+1\}$ includes the vertices associated with~$n$ pickup and delivery (p-d) pairs of customers as well as the vertices $\{0, 2n+1\}$, representing the initial and final depots locations. The set $P = \{1,2,\dots,n\}$ represents the pickup customers, while the set $D = \{n+1,\dots,2n\}$ represents the corresponding delivery customers. Each service~$i$ consists of a pickup customer $i \in P$ and a delivery customer $(n+i) \in D$. A positive demand $q_i > 0$ is associated with each pickup customer $i \in P$, and a negative demand, $q_{n+i} = -q_i$, is associated with the corresponding delivery customer $(n+i) \in D$. Each edge $(i,j)\in E$ represents the possibility of traveling from a vertex $i \in V$ to a vertex $j \in V$ with a distance cost $d_{ij}$.

A homogeneous fleet of $m$ vehicles with capacity limit $Q$ is available to perform the services.
Any vehicle arriving at a pickup vertex can collect all available load, or only a part of it. When a vehicle arrives at a delivery vertex, all load from this vehicle intended for this customer is delivered. As in previous works, we assume that $q_i \leq Q$ for all services $i$. Moreover, we impose in this paper a maximum travel distance $L$ for each vehicle.

The objective of the MPDPSL is to design a set of up to $m$ routes, starting and ending at the depot, with minimum travel distance, in such a way that the complete demand of each pickup and delivery is satisfied, the route distance limit is respected as well as vehicle capacities, and each pickup precedes its delivery in the same route. 

\begin{figure}[htbp]
\centering
\vspace*{0.1cm}
 \includegraphics[scale=0.43]{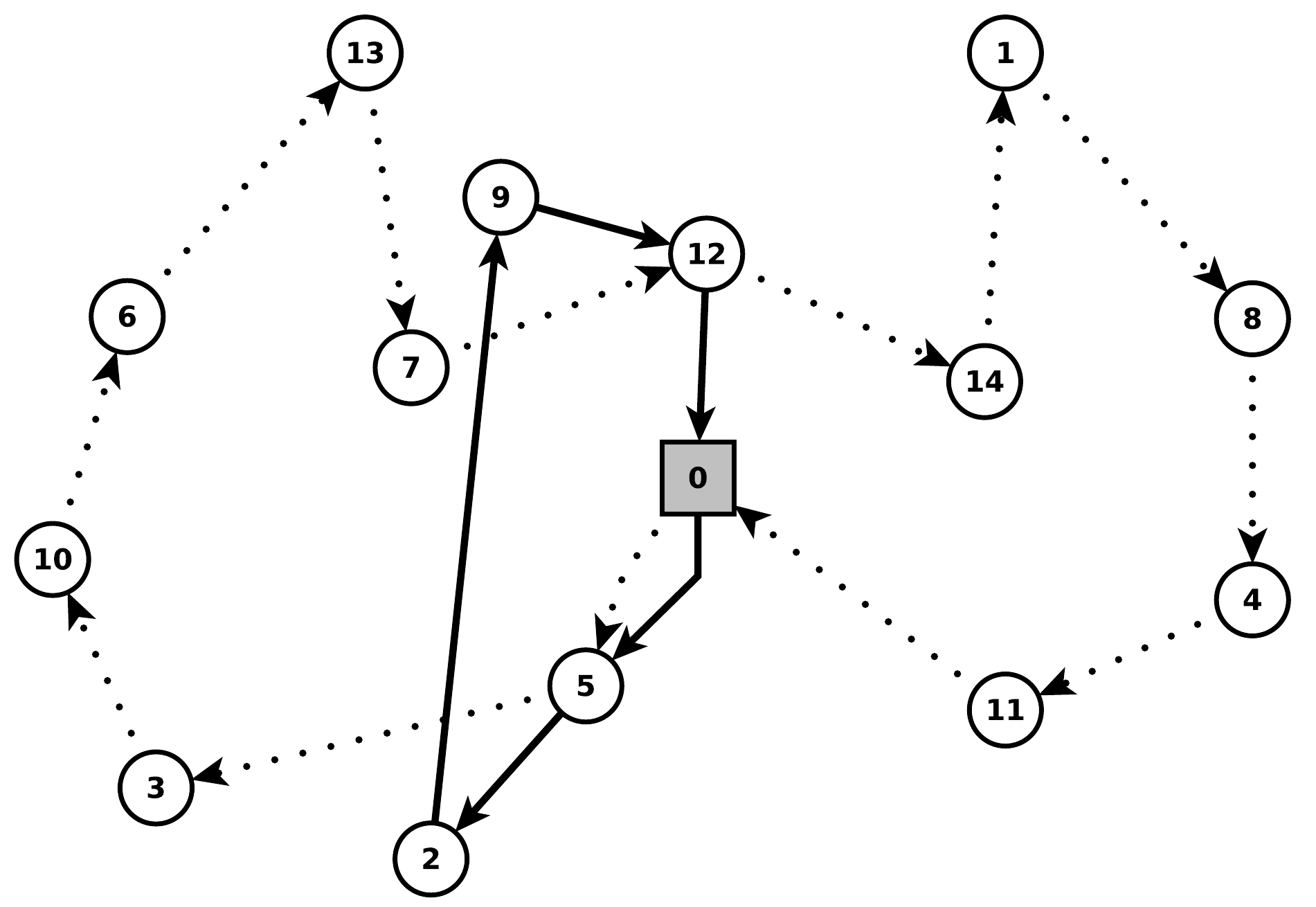}
\caption{\myblue{A feasible MPDPSL solution. The service to (2,9) is split among two routes.} \label{ex:PRVCEFU}}
\end{figure}

We adopt the same definition of a split load as in the SDVRP literature: a split load occurs when the demand of a customer is serviced by a larger number of trips than the minimum necessary. Figure \ref{ex:PRVCEFU} illustrates a feasible MPDPSL solution for an instance with seven p-d pairs served by two vehicles. \myblue{In this example, the service of the pair (5,12) is split among two routes, and each vehicle carries a fraction of the demand associated to this service.}
    
\begin{figure}[htbp]
\centering
 \includegraphics[scale=0.48]{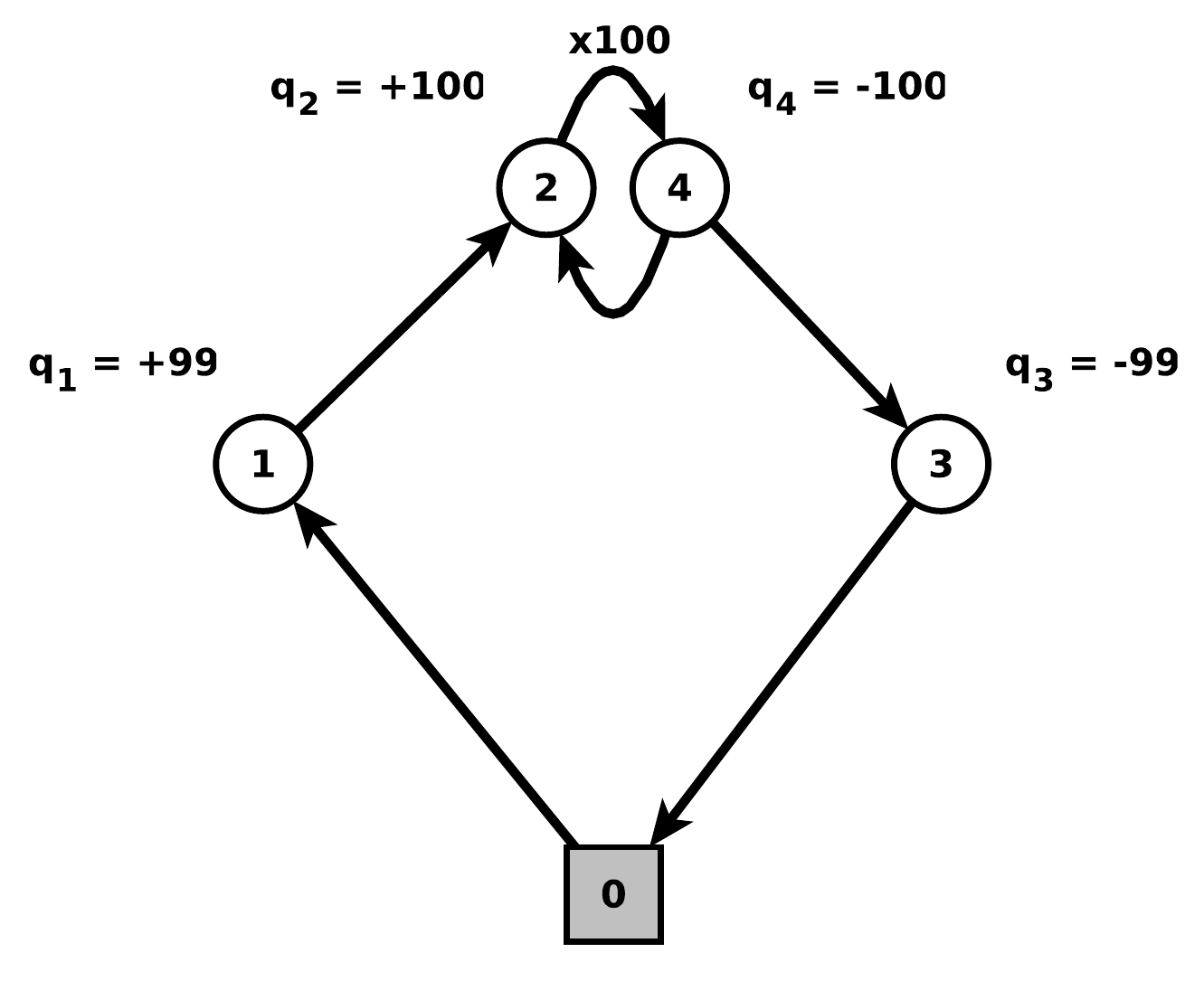}
\caption{Optimal MPDPSL solution for an instance with two p-d pairs, involving an arbitrary high number of visits to the same customer in a single route.\label{ex:case}}
\vspace*{-0.3cm}
\end{figure}

\myblue{Note that the previous example is not the only situation in which a split load can occur.} The MPDPSL is notably different from the SDVRP as more than one split load can occur \emph{in the same route} in an optimal solution. The solution size can also increase significantly and become, in some cases, exponential as a function of the input size. An illustrative example of such a situation is presented in Figure \ref{ex:case}.
This example includes two p-d pairs, (1,3) and (2,4), with one vehicle of capacity $Q=100$ and distance limit $L = \infty$. Customer $1$ wishes to transfer $99$ load units to customer~$3$, while customer $2$ wishes to transfer $100$ load units to customer $4$. The distance between customers $2$ and $4$ is small ($d_{24} = d_{42} = \epsilon$).

 The optimal solution for this problem instance contains a single tour, which
\begin{itemize}[nosep]
\item[--] visits customer $1$ and collects its full load (99 units), then
\item[--] visits customer $2$ to collect one unit of load and delivers it to customer $4$;
\item[--] repeats the previous operation $99$ additional times,
\item[--] delivers the load of customer $3$ ($99$ units), and returns to the depot.
\end{itemize}

This optimal solution performs $202$ customer visits instead of $4$ for the same problem without split loads. Of course, this is an extreme case of the classical MPDPSL, as in practical applications a base service time may be counted (e.g., as part of the travel distance), hence increasing~$d_{24}$. Nevertheless, a good heuristic should be able to find multiple split loads, since these situations can naturally occur.

\section{Related literature}
\label{sec:literature}

The PDP and SDVRP have been the focus of extensive work, counting hundreds of scientific articles. We refer to \cite{Archetti2012a,Battarra2014a,Doerner2014} and \cite{Irnich2014a} for detailed surveys. Among the current state-of-the-art methods for PDP, we can highlight, in particular, the adaptive large neighborhood search (ALNS) of \cite{Ropke06}, which iteratively improves an incumbent solution by means of repeated destruction and reconstruction steps, and the hybrid genetic algorithm of \cite{Nagata2011} based on a selective route exchange crossover with efficient local searches, producing to this date the best results for PDP instances with time constraints. ALNS has been subsequently extended to a wide range of VRP variants with great success.

For the SDVRP, the current state-of-the-art results are obtained by the multi-start ILS-RVND metaheuristic of \cite{Silva15}. This method exploits a wide collection of construction techniques and neighborhoods for solution improvement, as well as a perturbation strategy which operates multiple random \emph{k-Split} moves. Finally, \cite{Chen2017} \myblue{proposed} an \emph{a-priori Split strategy}, in which customer's demands are split in advance, and a capacitated VRP (CVRP) solver is subsequently used. This simple approach leads to solutions of fair quality and leverage decades of CVRP research.

In contrast, very few articles have considered the combination of both attributes in a single problem. To our knowledge, \cite{Mitra05} first considered a problem related to the PDPSL, but with simultaneous pickups and deliveries instead of one-to-one requests. The objective seeks to minimize the fleet size and then the distance. The authors propose a mixed integer programming (MIP) formulation for this problem and a route construction heuristic, which firstly determines the minimum number of vehicles required, and then builds routes based on a cheapest insertion criterion. An additional MIP formulation and an extension of this heuristic, using parallel clustering, are proposed in \cite{Mitra08}.

\cite{Nowak08} evaluate the benefits of allowing split loads in the one-to-one PDP, hence defining the PDPSL. The objective of the PDPSL is to find a single route with minimum cost, fulfilling the required demand. A heuristic based on simulated annealing and tabu search is developed and random large-scale instances are created. The authors observe that the benefits of split loads are closely linked to three characteristics of the instances: the load size, the cost associated with the pickup or delivery, and the percentage of loads which have pickup and delivery locations in common. They also show that, for a given set of origins and destinations, the greatest benefits are observed when the load size is greater than half the capacity of the vehicle. A variant of the problem addressed in \cite{Nowak08} can be found in \cite{Thangiah07}, with additional time-window constraints. This work describes an algorithm that inserts shipments into vehicles using multiple-insertion heuristics for static and real-
time~test~cases.

\cite{Nowak09} perform an additional empirical analysis of the heuristic presented in \mbox{\cite{Nowak08}}. The authors note that when demands are between 51\% and 60\% of the capacity of the vehicle, up to 30\% transportation costs can be saved. The potential savings due to split loads depends on the percentage of loads to be collected or delivered in a common location, and the average distance from an origin to a destination relative to the distance from origin to origin and destination to destination.

\cite{Sahin13} consider the PDPSL with multiple vehicles and distance constraints and formally define the MPDPSL. The authors develop a heuristic based on tabu search and simulated annealing. The initial solution is built using a variant of the savings algorithm by \cite{Clarke64}, and then improved by local searches based on swap and insert/split neighborhoods. The simulated annealing is then used in combination with a tabu list to control move acceptance. Experiments are conducted on the instances from \cite{Nowak08}, as well as adapted instances from \cite{Ropke06}. However, since no distance limits are imposed to the vehicles, it is always better to use a single route, such that these instances cannot be viewed as multi-vehicle test cases.

Finally, \cite{Sahin13} and \cite{Oncan2011} also introduce an integer programming model \myblue{and branch-and-cut algorithm} for the MPDPSL, allowing to solve some problem instances with seven \emph{p-d} pairs. \myblue{This method, however, does not necessarily produce an optimal solution for the problem (e.g., in cases similar to Figure \ref{ex:case}), as the model allows at most one visit per route for each p-d pair. More generally, no compact edge-flow formulation with a strongly polynomial number of variables can be built for the MPDPSL, since the number of customer visits of an optimal solution may grow exponentially with the size of the instance. To overcome this issue, we propose a branch-and-price algorithm, such that the inherent complexity related to multiple split deliveries within the same route is relegated into the labeling algorithm used for column generation. This allows to generate the first known optimal solutions for the MPDPSL without any restriction on the number of visits.
}

\myblue{
\section{Exact Solution Approach}
\label{sec:exact}

This section first introduces a set partitioning formulation of the MPDPSL (Section \ref{sec:sp}), and then describes the column generation procedure (Section \ref{sec:cg}), and the branch-and-price algorithm (Section \ref{sec:bp}) designed to solve this problem to optimality.
}

\myblue{
\subsection{Mathematical Formulation}
\label{sec:sp}

The mathematical formulation used is an adaptation of the well-known set partitioning formulation, which is extensively used in successful exact approaches for vehicle routing problems. It considers the set of all feasible routes $\Omega$ and binary variables $\lambda_r$, representing whether route $r \in \Omega$ is used in the solution. In contrast with most other vehicle routing problems, set $\Omega$ naturally contains non-elementary routes, i.e., routes visiting vertices more than once due to split deliveries. 
In any route $r \in \Omega$, the precedence and capacity constraints should be respected, and the total amount serviced for any p-d pair must not exceed its demand.
\begin{align}
	\textrm{Minimize} \hspace*{0.7cm} \sum_{r \in \Omega}d_r \lambda_r & \label{eq:sp-obj} & \\
	\textrm{subject to} \hspace*{0.6cm} \sum_{r \in \Omega}\bar{q}_{ri} \lambda_r & = q_i & i \in P \label{eq:sp-c2} \\
	\lambda_r & \in \{0, 1\} & r \in \Omega. \label{eq:sp-c3}
\end{align}

The Objective Function \eqref{eq:sp-obj} minimizes the total cost of the solution. Given $\bar{q}_{ri}$ as the total amount of the p-d pair $(i, n + i)$ demand serviced by route $r$, Constraints \eqref{eq:sp-c2} guarantee that all demands of all p-d pairs are satisfied. Constraints \eqref{eq:sp-c3} are the variable bounds constraints.
Clearly, set $\Omega$ contains an exponential number of routes, and the above formulation cannot be solved by considering all variables. Therefore, a column generation approach is required to efficiently solve it, as presented in the next section.
}

\myblue{
\subsection{Column Generation}
\label{sec:cg}

Since the MPDPSL combines features of both pickup-and-delivery and split delivery problems, our Column Generation (CG) algorithm is first built upon an algorithm for the PDP with time windows (PDPTW), \myblue{as the one} presented in \cite{Ropke2009}, and then extended to consider split loads. The algorithm may generate routes considering all possible loads for all p-d pairs visited, under the condition that they respect the precedence constraints and the maximum travel distance $L$.

The CG starts with no routes and iteratively generates feasible ones by solving a pricing subproblem algorithm. Given that with no routes, the formulation of Section \ref{sec:sp} would become infeasible, we introduce one artificial variable for each constraint and solve a two-phase CG, following the same idea as the two-phase Simplex algorithm. At each iteration, the pricing subproblem must find one or more variables with negative reduced cost. Given $\beta_i$, the dual variables of Constraints \eqref{eq:sp-c2}, the reduced cost of a route can be calculated as $\bar{d}_r = d_r - \sum_{i \in P} \beta_{i} \bar{q}_{ri}$. The dual variable $\beta_i$ is only counted when the route visits a pickup vertex, and it is multiplied by the total amount of demand loaded in this pickup vertex. Therefore, we can write the reduced cost of an edge $(i, j) \in E, i \in V, j \in P$ that loads an amount $q$ of vertex $j$'s demand as $\bar{d}^{q}_{ij} = d_{ij} - \beta_j q$. On the other hand, the amount of demand unloaded on a delivery vertex does not imply any change on the route's reduced cost. For this reason, we define the reduced cost of an edge $(i, j) \in E, i \in V, j \in D$ simply as $\bar{d}_{ij} = d_{ij}$.

A partial path $(i, d, \cQ, q)$ is a non-elementary path that starts at the depot, visits a subset of vertices and ends in vertex $i$ servicing $q$ units of its demand with total travel distance $d$. Vector $\cQ$ contains the load of all opened p-d pairs. It also allows to know which delivery vertices must be visited in the future, and the current total load in the vehicle.

Note that the definition of partial paths allows infeasible routes with excess demand for some p-d pairs. This relaxation is used to facilitate the resolution of the pricing subproblem. Moreover, infeasible routes will anyway be excluded in any integer solution of the formulation thanks to Constraints \eqref{eq:sp-c2}.

The pricing subproblem is a \myblue{Resource-Constrained Shortest Path Problem} solved by a dynamic programming algorithm that works on a state-space graph $\cG = (\cV, \cE)$, with $\cV = \cV^P \cup \cV^D \cup \{(0, 0, 0^n, 0)\}$, where $\cV^P = \{(i, d, \cQ, q): \forall\cQ \in \Phi, \cQ[i] = q, 1 \leq q \leq q_i, \forall i \in P, 0 \leq d \leq L\}$, $\cV^D = \{(i, d, \cQ, 0): \forall\cQ \in \Phi, \cQ[i - n] = 0, \forall i \in D, 0 \leq d \leq L\}$, $(0, 0, 0^n, 0)$ represents the original depot vertex and $\Phi = \{\cQ \in \Nat^n: 0 \leq \cQ[i] \leq q_i, \forall i \in P, \sum_{i \in P} \cQ[i] \leq Q\}$. Furthermore $\cE = \{((j, d^\prime, \cQ^\prime, q^\prime), (i, d, \cQ, q)) : \forall (j, d^\prime, \cQ^\prime, q^\prime) \in \cE^{-1}(i, j, d, \cQ), \forall (j, i) \in E, \forall (i, d, \cQ, q) \in \cV, d_{ji} \leq d \leq L\}$, where $\cE^{-1}(j, i, d, \cQ) = \{ (j, d - d_{ji}, \cQ^\prime, q^\prime) : \forall \cQ^\prime \in \Phi \text{ s.t. } \cQ^\prime[k] = \cQ[k] \ \forall k \in P \backslash \{i\} \text{ and } \cQ^\prime[i] = 0 \text{ if } i \in P, \cQ^\prime[k] = \cQ[k] \ \forall k \in P \backslash \{i - n\} \text{ and } \cQ^\prime[i - n] > 0 \text{ if } i \in D, \text{ or } \cQ^\prime[k] = 0 \ \forall k \in P \text{ if } i = 0, 0 \leq q^\prime \leq q_j\}$. The recursion can then be written as:
\begin{equation}
f(i, d, \cQ, q) = \min_{(j, d^\prime, \cQ^\prime, q^\prime) \in \cE^{-1}(j, i, d, \cQ)} \{f(j, d^\prime, \cQ^\prime, q^\prime) + \bar{d}^q_{ji}\}, \forall (i, d, \cQ, q) \in \cV.
\end{equation}

In order to reduce the number of states, we use the following dominance rule. State $(i, d, \cQ, q)$ dominates state $(j, d^\prime, \cQ^\prime, q^\prime)$ iff (i) $i = j$, (ii) $d \leq d^\prime$ (iii) $f(i, d, \cQ, q) \leq f(j, d^\prime, \cQ^\prime, q^\prime)$ and (iv) $\cQ[k] \leq \cQ^\prime[k], \forall k \in P$. Note that (iv) assures the total load on partial path $(i, d, \cQ, q)$ to be less than the one on partial path $(j, d^\prime, \cQ^\prime, q^\prime)$. Moreover, this condition is only valid if the reduced costs respect the Delivery Triangle Inequality (DTI) \citep{Ropke2009}. An MPDPSL cost matrix is said to respect the DTI if $d_{ij} \leq d_{ik} + d_{kj}, \forall i, j \in V, k \in D$. If the original distances $d_{ij}$ respect the DTI, $\bar{d}^q_{ij}$ will also respect the DTI based on the definition \myblue{previously presented}.

We apply three additional techniques to improve the CG algorithm. First, we use the dual stabilization procedure proposed in \cite{Pessoa2010}. At each CG iteration, let $\beta^\prime$ be the dual solution of the previous iteration and let $\alpha \in [0, 1[$ be the dual stabilization parameter. The CG uses a composition of the current and the previous dual solution calculated as $\hat{\beta} = \alpha \beta^\prime + (1 - \alpha) \beta$. Parameter $\alpha$ starts with a positive value, and each time the pricing subproblem returns a route with positive reduced cost, $\alpha$ is reduced until it reaches zero, thus concluding the dual stabilization procedure.}

\myblue{Moreover, we use a succession of heuristic pricing algorithms to save computational effort. We first execute a version of the dynamic programming that limits the number of partial paths stored for each $i \in V, 0 \leq d \leq L$ by one. When this simple heuristic fails, we call the exact pricing relaxing Condition (iv) of the dominance rule and sequentially limiting the number of partial paths stored for each $i \in V, 0 \leq d \leq L$ by $\{ 3, 10, 100, \infty\}$. When no route is found with no limitation on the number of partial paths, we restore Condition (iv) and repeat the procedure.}

\myblue{Finally, we use pre-processing to identify forbidden extensions due to the travel distance limit. For each $0 \leq d \leq L$, we forbid an extension from vertex $i$ to vertex $j$ based on the following rules: (i) if $i \in P, j \in P, d + d_{ij} + d_{j(n+i)} + d_{(n+i)(n+j)} + d_{(n+j)0} > L$ and $d + d_{ij} + d_{j(n+j)} + d_{(n+j)(n+i)} + d_{(n+i)0} > L$, (ii) if $i \in P, j \in D$ and $d + d_{ij} + d_{j(n+i)} + d_{(n+i)0} > L$, (iii) if $i \in D, j \in P$ and $d + d_{ij} + d_{j(n+j)} + d_{(n+j)0} > L$, and (iv) if $i \in D, j \in D$ and $d + d_{ij} + d_{j0} > L$. This is an extension of the rules created by \cite{Dumas1991} for the PDPTW, adapted to consider the travel distance limit.}

\myblue{
\subsection{Branch-and-Price}
\label{sec:bp}

Branch-and-Price (B\&P) is the name given when a CG algorithm is used on each node of a branch-and-bound procedure to obtain an optimal integer solution. After the solution of a node, a fractional variable (or a set of variables) is chosen and two or more branches are created by introducing constraints to cut the fractional value. When solving a B\&P, the branching rules must be carefully chosen, otherwise the CG algorithm may price the same variable again. For this reason, the branching rules used within a B\&P are usually on ``original variables'', i.e., variables from an edge-flow formulation.

Our B\&P uses four branching rules. The first one considers the number of vehicles used in the solution. It can be calculated as $\sum_{r \in \Omega} \lambda_r$. The second branching rule is done on the degree of each vertex. Given $\bar{a}_{ri}$, the number of times route $r$ visits pickup $i \in P$, the degree can be calculated as $\sum_{e \in \Omega} \bar{a}_{ri}\lambda_r, \forall i \in P$. The third one is done on the edges of the original graph. Given $\bar{b}_{re}$, the number of times route $r$ traverses edge $e \in E$, regardless of the load, it can be calculated as $\sum_{r \in \Omega} \bar{b}_{re}\lambda_r, \forall e = (i, j) \in E$. Finally, the algorithm also considers a branching rule on the number of times an edge $e = (i, j)$ is traversed when loading (or unloading) $\bar{q}_i$ units of demand on vertex $i \in V$ and loading (or unloading) $\bar{q}_j$ units of demand on vertex $j \in V$.

Each constraint added to the master formulation generates a new dual variable, which must be considered by the pricing subproblem to calculate the reduced cost of the routes. It is not a difficult task to associate the new dual variables to the edges' reduced cost. However, while the first two branching rules do not violate the DTI, the last two may change the reduced cost matrix in this sense. To overcome this issue, we apply the fix proposed by \cite{Ropke2009}.}
\myblue{The remaining components of the B\&P are classical in the routing literature.
The branching rules are used in order, at each iteration the algorithm chooses the most fractional value, and the nodes are explored using the best-first strategy.}

\myblue{\section{Large Neighborhood-Based Metaheuristic}}
\label{Metodologia}

As noted in Section \ref{sec:literature}, notably few heuristics have been designed for the MPDPSL, and these methods were evaluated on benchmark instances that only require the use of a single vehicle. \myblue{Moreover, even the sophisticated B\&P algorithm described in Section \ref{sec:exact} is limited to instances of small and medium sizes.} To solve larger test cases, 
we design a simple and efficient metaheuristic for the MPDPSL, based on a new exponential-size neighborhood, and investigate its performance on distance-constrained benchmark instances that require multiple vehicles.

The proposed ILS--PDSL is built around a very simple search scheme which consists, as in the classical ILS metaheuristic, of iteratively improving a solution via neighborhood searches until reaching a local minimum, and then applying a perturbation operator to escape. This process is repeated until a termination criterion (a time limit in our case) is attained. The general pseudo-code of the method is displayed in Algorithm~\ref{ILS--PDSL}.

\begin{algorithm}[htbp]
\small
\SetKw{retorne}{Return}

$s_\textsc{best} \gets \varnothing$ \;
$s \leftarrow$ \myFont{Construct Initial Solution} \; 

\While{\myFont{Termination Criterion is not attained}}
{
	$s \leftarrow$  \myFont{RVND}$(s)$  \tcp*{\myFont{Solution improvement via a RVND}}
        $s \leftarrow$  \myFont{RCSP-insertion}$(s)$   \tcp*{\myFont{Large neighborhood search}}

\If{$s_\textsc{best} = \varnothing$ \textbf{\emph{or}} \myFont{Cost}$(s)<$ \myFont{Cost}$(s_\textsc{best})$}
{
$s_\textsc{best} \gets s$    \tcp*{\myFont{Saving the best solution}}
}

$s \leftarrow$  \myFont{Perturbation}($s_\textsc{best}$)  \tcp*{\myFont{Perturbation to prepare for next iteration}}

}
\retorne{$s_\textsc{best}$} \;

\caption{ILS--PDSL\label{ILS--PDSL}}
\end{algorithm}

Despite its apparent simplicity, the proposed metaheuristic differs from traditional ILS due to the nature of the neighborhoods used for solution improvement. Instead of relying on local search, it exploits a two-phase improvement method. The first phase is a randomized variable neighborhood descent (RVND), which explores a variety of neighborhoods in random order, and the second phase is a search in a new exponential-size neighborhood, \myblue{called resource-constrained shortest path insertion (RCSP-insertion),} which allows to optimally split and re-insert each pickup and delivery. Finally, our perturbation operator is always applied on the current best solution in an effort to direct the search on more promising regions of the search space.

The remainder of this section details each component of the algorithm, starting with the construction of the initial solution (Section~\ref{initialsol}), the RVND procedure (Section~\ref{modrvnd}), the \myblue{RCSP-insertion operator} (Section~\ref{largeneighbor}), and finally the perturbation mechanism (Section~\ref{modvns}). With the exception of the exponential-size neighborhood search performed in RCSP-insertion, these procedures are relatively simple and classic, leading to a high-performance algorithm which can be easily reproduced.

\subsection{Initial solution}
\label{initialsol}

The initial solution $s$ is produced by a greedy constructive heuristic.
Iteratively, this heuristic computes for each pickup customer $i$ its best feasible insertion position, with minimum increase of distance. The pickup customer $i$ with the shortest distance increase is inserted at each iteration, and the corresponding delivery $(n+i)$ is added in its best feasible position after $i$. At this stage, the method only considers the insertion of full deliveries. Moreover, only feasible insertions in terms of load capacity and distance constraints are enumerated, and a new route is created if no such position exists.

\subsection{Randomized variable neighborhood descent}
\label{modrvnd}

We first recall the concept of \emph{block} \citep{Cassani04}, which is needed to describe some neighborhoods. A block $B_i$ is defined as a sequence of consecutive visits that starts at a pickup customer $i$ and ends at the corresponding delivery customer $(n + i)$. A block $B_i$ is a simple block if there is no customer between $i$ and $(n + i)$. A block $B_i$ is a compound block when there is at least one block $B_j \in B_i$ such that $\Pi(i) < \Pi(j) < \Pi(n + j) < \Pi(n + i)$, where $ \Pi(i)$ is the position of the customer $i$ in the route. It is noteworthy that a compound block cannot contain a pickup customer without its corresponding delivery customer and vice versa.

As in the RVND of \cite{souza10} and \cite{subra10}, there is no predefined order for the neighborhoods, that is, before every execution of the local search, a new neighborhood order is randomly chosen. Each neighborhood is defined relatively to one type of move, which can be applied on different p-d pairs and routes. Each neighborhood is evaluated exhaustively, considering the moves in random order of p-d pairs, and applying the first improving move.
After each improvement, the search restarts from the first neighborhood structure. Otherwise, the search continues on the next neighborhood structure and finishes when all the neighborhoods have been examined without success.
Our RVND uses simple extensions of known enumerative neighborhoods for vehicle routing and pickup-and-delivery problems, which are listed in the following. Neighborhoods $N^{(1)}$, $N^{(2)}$, $N^{(5)}$, and $N^{(6)}$ can be traced back to \cite{Cassani04}.

\paragraph{Intra-route neighborhood structures:}

\begin{description}[nosep]
\item[$N^{(1)}$] -- \emph{PairSwap} considers two pairs of customers $(i,n+i)$ and $(j,n+j)$ and swaps the pickup customer $i$ with the pickup customer $j$, as well as the delivery customer $(n+i)$ with the delivery customer $(n+j)$.
\item[$N^{(2)}$] -- \emph{PairShift} considers a pair of customers $(i,n+i)$ and relocates the pickup $i$ in a position of the interval $[\Pi(i)-\Delta,\Pi(i)+\Delta]$ and the delivery $(n+i)$ in a position of the interval $[\Pi(i)+1,\Pi(i)+\Delta]$. Parameter $\Delta$ limits the size of the neighborhood~(see Section \ref{Resultados}).
\item[$N^{(3)}$] -- \emph{PickShift} relocates a pickup customer $i$ in another position before the delivery customer~$(n+i)$.
\item[$N^{(4)}$] -- \emph{DelShift} relocates a delivery customer $(n+i)$ in another position after the pickup customer $i$.
\item[$N^{(5)}$] -- \emph{BlockSwap} swaps a block $B_i$ with another block $B_j$. 
\item[$N^{(6)}$] -- \emph{BlockShift} relocates a block $B_i$ in another position.
\end{description}

\paragraph{Inter-route neighborhood structures:}

\begin{description}[nosep]
\item[$N^{(7)}$] -- \emph{InterPairSwap} selects a pair of customers $(i,n+i)$ from a route $r_1$ and another pair $(j,n+j)$ from a route $r_2$ and swaps the pickup customer $i$ with the pickup customer $j$. The delivery customer $(n+i)$ is swapped with the delivery customer $(n+j)$.
\item[$N^{(8)}$] -- \emph{InterPairShift} takes a pair of customers $(i,n+i)$ from a route $r_1$ and transfer this pair to a route $r_2$. After defining $\Pi(i)$ in $r_2$, the delivery customer is inserted in a position of the interval $[\Pi(i)+1,\Pi(i)+\Delta]$.
\item[$N^{(9)}$] -- \emph{InterBlockSwap} selects a block $B_i$ from a route $r_1$ and another block $B_j$ from a route $r_2$ and swaps them. 
\item[$N^{(10)}$] -- \emph{InterBlockShift} transfers a block $B_i$ from a route $r_1$ to a route $r_2$. \\ \vspace*{-0.2cm}
\end{description}

\noindent
Finally, we rely on the following theorem to perform a post-optimization after each local search:

\begin{theorem}[\citealt{Sahin13}]
If the distance matrix satisfies the triangle inequality, then there exists an optimal solution of the MPDPSL such that between each visit to a pickup customer $i$ and its corresponding delivery $(n+i)$ no other pickups or deliveries of this same p-d pair occur.
\label{theo}
\end{theorem}

As such, we scan the solution and search for visits to the same p-d pair $(i,n+i)$ appearing in the order $i \rightarrow i \rightarrow n+i \rightarrow n+i$ in a route (with possible visits to other customers in-between). If this situation occurs, the two visits can be merged as one single visit while maintaining feasibility and improving the total distance. There are $2 \times 2$ possibilities for insertion of the merged p-d in place of the previous services, and the best one in terms of distance is chosen.

\myblue{\subsection{RCSP-insertion neighborhood}}
\label{largeneighbor}

The improvement procedure of the previous section relies on the enumeration of many possible moves to produce improved solutions. However, we know that MPDPSL solutions can include an arbitrarily large number of visits to the same p-d pair (as illustrated in Figure \ref{ex:case}). Enumerating all possible combinations of splits and placements of visits would take an exponential time. For this reason, previous methods adopted strategies which limit the number of split loads \citep{Nowak08,Sahin13}. To address this issue, we propose a larger (exponential-size) neighborhood, which seeks to optimize the split loads and can be efficiently explored via dynamic programming. 

In the proposed RCSP-insertion neighborhood, the problem of finding the best reinsertion of each pickup and delivery pair, with possible split loads, is addressed as a resource-constrained shortest path problem (RCSP) in a directed acyclic graph followed by a knapsack problem. This optimization is conducted once for each p-d pair, considering the pairs in random order. For each p-d pair ($x$, $n+x$), the method works as follows:
\begin{itemize}[nosep]
\item[--] Remove all occurrences of $x$ and $n+x$ from all routes.
\item[--] Phase 1: For each route $\sigma$,
evaluate the possible insertions and combinations of insertions of the p-d pair $(x,n+x)$ via dynamic programming (RCSP), therefore characterizing all non-dominated trade-offs between the extra travel distance and the quantity of load picked-up from $x$ and delivered to $n+x$.
\item[--] Phase 2: Based on the known trade-offs (labels) for each route, find the best combination of insertions in all routes in order to fulfill the total demand $q_x$. This selection can be done by solving a variant of the knapsack problem.
\end{itemize}

\begin{figure}[H]
\begin{center}
	\includegraphics[scale=0.75]{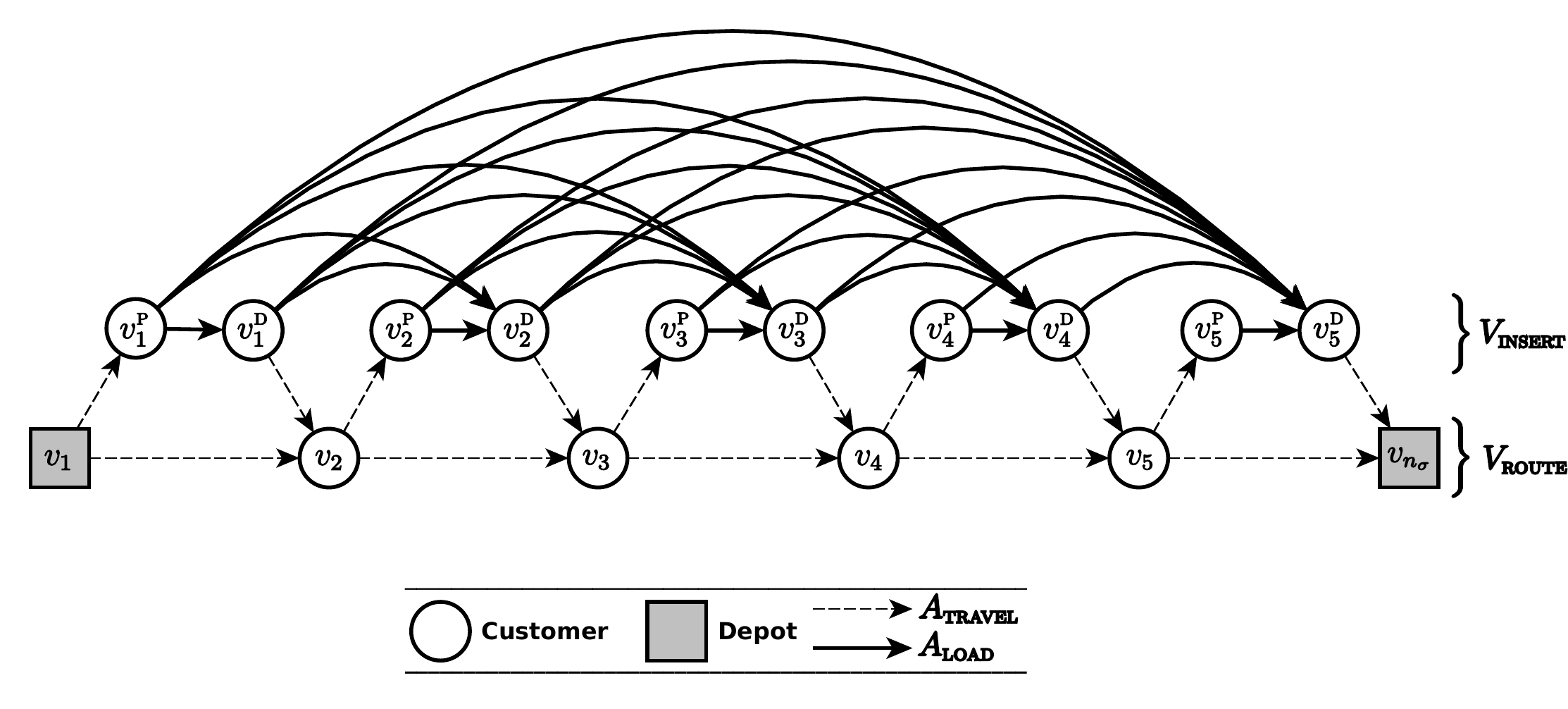}
	\caption{Auxiliary graph $H$ for a route containing $n(\sigma) = 6$ visits \label{rcsp}}
\vspace*{-0.3cm}
\end{center}
\end{figure}

\paragraph{Phase 1: Evaluation of non-dominated insertions for each route.}~\myblue{Consider a route $\sigma = (\sigma_1, \dots, \sigma_{n(\sigma)})$, in which each element represents a visit to a depot, pickup or delivery node.} This phase aims to evaluate the minimum additional distance incurred when inserting visits to the p-d pair $(x,n+x)$ in the route $\sigma$, in order to service any demand quantity $q$ in the interval $[0,q_x]$. Trade-offs between distance and delivery quantity can be found by solving a resource-constrained shortest path problem in a directed acyclic graph \mbox{$H = (V',A)$}, illustrated in Figure \ref{rcsp} \myblue{and defined in the following}.

\noindent
\textbf{The node set $V'$} is divided into two groups of nodes, $V' = V_\textsc{route} \cup V_\textsc{insert}$:
\begin{itemize}[nosep]
\item[--]  $V_\textsc{route} = \{v_1,\dots,v_{n(\sigma)}\}$ contains one node per (depot or customer) visit in the route.
\item[--]  $V_\textsc{insert} = \{v^\textsc{p}_1, v^\textsc{d}_1, \dots, v^\textsc{p}_{n(\sigma)-1}, v^\textsc{d}_{n(\sigma)-1}\}$ contains a pair of nodes $(v^\textsc{p}_i, v^\textsc{d}_i)$ between each node pair $(v_i,v_{i+1})$. The nodes $v^\textsc{p}_i$ represent possible pickups at $x$, and the nodes $v^\textsc{d}_i$ represent possible deliveries at $n+x$.
\end{itemize}
The total number of nodes in the graph is $|V'| = 3 \times n(\sigma) - 2$.\\

\noindent
\textbf{The arc set $A$} is also divided into two sets $A = A_\textsc{travel} \cup A_\textsc{load}$.
Each arc is~characterized by a distance $\delta^\textsc{dist}_a$ and a delivered load $\delta^\textsc{load}_a$.
The arcs in $A_\textsc{travel}$ (dashed arrows in Figure \ref{rcsp}) can either connect successive visits in $V_\textsc{route}$, or connect a visit $v_i$ with its candidate pickup~$v^\textsc{p}_{i}$, or connect a candidate delivery $v^\textsc{d}_{i}$ with the next visit $v_{i+1}$. Each such arc $a \in A_\textsc{travel}$ represents a pure vehicle relocation without any load destined for customer $x$, such that $\delta^\textsc{load}_a = 0$, and the associated distance is:
$$
\renewcommand{\baselinestretch}{1.2}
\delta^\textsc{dist}_a  =
\begin{cases} 
 d_{\sigma_{i},\sigma_{i+1}}& \text{ if }  a=(v_{i},v_{i+1}) \\
 d_{\sigma_{i},x} &\text{ if }  a=(v_{i},v^\textsc{p}_{i}) \\
 d_{n+x,\sigma_{i+1}} &\text{ if }  a=(v^\textsc{d}_{i},v_{i+1}).
\end{cases}$$
Finally, the arcs in $A_\textsc{load}$ (solid arrows in Figure \ref{rcsp}) correspond to trips which carry some load of $x$. The following cases should be distinguished.
\begin{itemize}
\item[--] \textbf{Direct arc:} $a = (v^\textsc{p}_{i},v^\textsc{d}_{i})$. This arc corresponds to a direct travel between $x$ and $n+x$. It is characterized by a distance $\delta^\textsc{dist}_a = d_{x,n+x}$ and a load  $\delta^\textsc{load}_a = Q - \sum\limits_{k=1}^i q_{\sigma_k}$, which corresponds to the \myblue{free} capacity \myblue{in} the vehicle after client $\sigma_i$.

\item[--] \textbf{Indirect pickup--delivery arc:} $a = (v^\textsc{p}_{i},v^\textsc{d}_{j})$ with $i < j$. This arc corresponds to a trip segment starting at the pickup location $x$, serving the locations $(\sigma_{i+1}, \sigma_{i+2}, \dots, \sigma_j)$, and ending at the delivery location $n+x$. Following the same principles as previously, \vspace*{-0.3cm}
$$\begin{cases} 
\delta^\textsc{dist}_a = d_{x,\sigma_{i+1}} + \displaystyle\sum_{k=i+1}^{j-1} d_{\sigma_k \sigma_{k+1}} + d_{\sigma_{j},n+x}, \text{ and } \\
\delta^\textsc{load}_a = \min\limits_{l \in \{i,\dots,j\}} \left(Q - \displaystyle\sum_{k=1}^l q_{\sigma_k} \right).
\end{cases}$$
\myblue{
In this equation, $\delta^\textsc{load}_a$ represents the smallest amount of free capacity in the vehicle at any point of the trip between $\sigma_{i}$ and $\sigma_j$.
}

\item[--] \textbf{Indirect delivery-delivery arc:} $a = (v^\textsc{d}_{i},v^\textsc{d}_{j})$ with $i < j$. This arc corresponds to a trip segment starting at the delivery location $n+x$, returning to the pickup location $x$, serving the locations $(\sigma_{i+1}, \sigma_{i+2}, \dots, \sigma_j)$, and ending at the delivery location $n+x$.  As such, \vspace*{-0.3cm}
$$\begin{cases} 
\delta^\textsc{dist}_a =  d_{x,n+x} + d_{x,\sigma_{i+1}} + \displaystyle\sum_{k=i+1}^{j-1} d_{\sigma_k \sigma_{k+1}} +  d_{\sigma_{j},n+x}, \text{ and } \\
\delta^\textsc{load}_a = \min\limits_{l \in \{i,\dots,j\}} \left( Q -  \displaystyle\sum_{k=1}^l q_{\sigma_k} \right).
\end{cases}$$
\end{itemize}

After the construction of the graph $H$, the RCSP between $v_1$ and $v_{n(\sigma)}$ is obtained by means of a simple variant of Bellman's algorithm. The algorithm computes for each vertex $v \in V'$, in topological order, a set of labels \mbox{$S_v = \{ s_{v k} \ | \ k \in \{1,\dots,|S_v| \} \}$} in which each label $s_{v k} = (s_{v k}^\textsc{dist}, s_{v k}^\textsc{load})$ is characterized by a distance $s_{v k}^\textsc{dist}$ and a load $s_{v k}^\textsc{load}$ transferred from $x$ to $n+x$.
Starting at the depot with $\mathcal{S}_{v_1} = \{(0,0)\}$, the labels are iteratively propagated as follows:  \vspace*{-0.3cm}
\begin{equation}
\begin{aligned}
&\text{ for } v \in (v^\textsc{p}_1,v^\textsc{d}_1,v_2,v^\textsc{p}_2,v^\textsc{d}_2,\dots,v_{n(\sigma)}), \\
&\hspace*{3cm}\mathcal{S}_{v} = \bigcup_{w|(w,v) \in A} ~ \bigcup_{s_{w i} \in \mathcal{S}_{w}} \{(s^{\textsc{dist}}_{w k} + \delta^\textsc{dist}_{(w,v)} , s^{\textsc{load}}_{w k} + \delta^\textsc{load}_{(w,v)} )\}.
\end{aligned}
 \label{pd_labels}
\end{equation}

Non-dominated labels are eliminated at each step.
A label $s_{v k}$ is dominated by a label $s_{v k'}$ if $s^\textsc{dist}_{v k} \geq  s^\textsc{dist}_{v k'}$ and $\min\{s^\textsc{load}_{v k},q_x\}  \leq  \min\{s^\textsc{load}_{v k'},q_x\}$.
Moreover, a completion bound is used to eliminate additional labels: for any $v_i \in V_\textsc{route}$, any label $s_{v_i k}$ that covers the total demand of the client $x$ (such that $s^\textsc{load}_{v_i k} \geq q_x$) leads to a distance bound of $s^\textsc{dist}_{v_i k} + \sum\limits_{k=i}^{n_\sigma-1} d_{k,k+1}$. The best distance bound is updated during the search, and any label whose distance exceeds this bound can be pruned.

For each route $\sigma$, the set of non-dominated labels $S(\sigma) = S_{v_{n(\sigma)}}$ is stored at the end of the algorithm. For single-vehicle problem instances, the best combination of insertions of visits to the p-d pair $(x,n+x)$ corresponds to the single non-dominated label $s \in S(\sigma)$ such that $s^\textsc{load} \geq q_x$. In cases involving multiple vehicles, the best visits for the p-d pair $(x,n+x)$ can be distributed into multiple routes. As described in the following, the best combination of insertions can be found by solving a knapsack problem based on the labels $S(\sigma)$ for each route~$\sigma$.

\paragraph{Phase 2: Combination of insertions in multiple vehicles.}
In the presence of multiple vehicles, the algorithm searches for a  good combination of insertions in different routes in order to cover the total demand. This problem can be formulated as a knapsack problem with an additional constraint that limits the selection to one label at most in each route. Let $C_\sigma$ be the distance of a route $\sigma$ before the insertion of any visit to the p-d pair $(x,n+x)$. Each label $s_{\sigma j} \in S(\sigma)$ corresponds to a detour cost of $s^\textsc{dist}_{\sigma j} - C_\sigma$, to deliver a load quantity $s^\textsc{load}_{\sigma j}$ from the pickup $x$ to the delivery $n+x$. We thus define a binary decision variable~$y_{\sigma j}$, equal to 1 if and only if the label $s_{\sigma j}$ is selected. This leads to the optimization problem of Equations (\ref{modelstart}--\ref{modelend}).
 \begin{align}
 \min \hspace*{0.3cm}& \displaystyle\sum_{\sigma \in \mathcal{R}} \displaystyle\sum_{s_{\sigma j}  \in S(\sigma)} \left( s^\textsc{dist}_{\sigma j} - C_\sigma \right)  y_{\sigma j} \label{modelstart} \\
 & \displaystyle\sum_{\sigma \in \mathcal{R}} \displaystyle\sum_{s_{\sigma j}  \in S(\sigma)}  s^\textsc{load}_{\sigma j}   y_{\sigma j} \geq q_x \\
 & \displaystyle\sum_{s_{\sigma j}  \in S(\sigma)}  y_{\sigma j} \leq 1 & \sigma \in \mathcal{R} \\ 
 &  y_{\sigma j} \in \{0,1\} & \sigma \in \mathcal{R}, \ s_{\sigma j}  \in S(\sigma) \label{modelend}
\end{align}

This formulation is identical to the one used in \cite{Boudia07} for the SDVRP. At this stage, the challenges specific to the MPDPSL have already been relegated~to the determination of the labels $(s^\textsc{dist}_{\sigma j}$,$s^\textsc{load}_{\sigma j})$; a task which could not be done by inspection in $O(n)$, but instead required a pseudo-polynomial search algorithm (Phase~1) to produce non-dominated pairs of insertion positions ---as well as combinations of insertion positions--- within each route.

To solve Equations (\ref{modelstart}--\ref{modelend}), we tested different exact techniques, either based on dynamic programming or integer programming. In our experiments, these methods led to a significant computational-time overhead. Similarly to \cite{Boudia07}, we thus opted for a heuristic resolution, using a greedy heuristic which iteratively selects the label $s_{\sigma j}$ with maximum ratio \myblue{$s_{\sigma j}^\textsc{load}/(s_{\sigma j}^\textsc{dist} - C_\sigma)$}. In our experiments, this heuristic matches in 69\% of the cases the optimal result. Finally, the best visit insertions are performed, forming the new incumbent solution in the algorithm.

\subsection{Perturbation mechanism}
\label{modvns}

The last component of ILS--PDSL, the perturbation mechanism, is designed to escape from the local minimums of the previous neighborhood improvement procedures. It relocates $n_\textsc{pert}$ random p-d pairs from their original routes to new random positions, inserting both pickup and deliveries consecutively.
The number of pairs $n_\textsc{pert}$ to be relocated, which determines the strength of the perturbation, is randomly selected in $\{1,2,...,p_\textsc{max} \}$  with uniform distribution. As such, $p_\textsc{max}$ is a method parameter which establishes a maximum limit on the impact of the perturbation.

\section{Computational results}
\label{Resultados}

Our computational experiments have been conducted on the two existing sets of PDPSL instances from previous literature, as well as new MPDPSL instances. The first set originates from \cite{Nowak08}, and the second from \cite{Sahin13}. These instances were generated in such a way that each load occupies 51\% to 60\% of the capacity of the vehicle. In those conditions, \cite{Nowak08} noticed that the savings related to split loads tend to be the greatest.

The first set contains three subsets of 15 instances each, with 75, 100 and 125 pickup and delivery pairs. In each instance, the pickups can occur in only five different locations, and each subset has a different number of delivery locations: 15, 20 and 25 delivery locations, respectively.
The second set was derived from the instances of \cite{Ropke06}. It contains four subsets of 12 instances each, with 50, 100, 250 and 500 pickup and delivery pairs. In this set, both pickup and delivery locations are randomly generated, such that coincident service locations are unlikely.

\myblue{Both exact and heuristic approaches were developed in C++. ILS--PDSL was implemented using} OptFrame \citep{Coelho11}, a computational framework for the development of efficient heuristic algorithms for combinatorial optimization problems. Each test was executed on a single thread, the ILS--PDSL on an Intel Core 2 Quad 2.4~GHz with 4~GB of RAM, \myblue{and the B\&P on an Intel Core i7-3960X 3.3~GHz with 64~GB of RAM}. We compare the performance of ILS--PDSL with that of the ``TESA'' algorithm of \cite{Nowak08}, and the ``TABU'' search of \cite{Sahin13}. Our computer is nearly identical to the one used in \cite{Sahin13}: an Intel Core 2 Quad 2.33 GHz with 3.46 GB of RAM. Moreover, the CPU time of \cite{Nowak08} has been scaled in \cite{Sahin13} to take into account the speed difference between their respective computers. 

ILS--PDSL uses three main parameters: the strength of the perturbation operator $p_\textsc{max}$, the range of insertions $\Delta$ considered in the \emph{PairShift} neighborhoods, and the stopping criterion~$T_\textsc{max}$. The first two parameters have been calibrated in preliminary analysis, considering values of $p_\textsc{max} \in \{1,\dots,10\}$ and $\Delta \in \{1,\dots,10\}$, and the configuration $p_\textsc{max} = 3$ and $\Delta = 5$ led to good results. Finally, the stopping criterion $T_\textsc{max}$ has been set to be identical to that of the TABU search of \cite{Sahin13}, for each group of instances, in order to compare with previous authors in similar CPU time.

Depending on the instance set, previous authors have either reported results on a single run, or best results over multiple runs. Both measures tend to be influenced by the variance of the performance of an algorithm over different runs with different seeds. We thus opted to report the average solution quality over several runs, which is a better estimate of the average behavior of an algorithm.
In the following, we will report solution values and their ``Gap(\%)'' for each instance. Let $z$ be the solution value of the proposed method, and $z_\textsc{bks}$ be the best known solution (BKS) ever found in previous literature for this instance (possibly over multiple runs, with different algorithms and parameter settings), then Gap($\%$) $= 100 \times (z  - z_\textsc{bks})/z_\textsc{bks}$.

\pagebreak

\subsection{Metaheuristic -- Performance evaluation on PDPSL instances}
\label{test11}

\myblue{We first evaluate the performance of the ILS--PDSL.
For this purpose, we establish a comparison with previous metaheuristics available in the literature, which were tested on PDPSL instances.}

\paragraph{Instances from \cite{Nowak08}.}
\cite{Nowak08} and \cite{Sahin13} reported the solution quality of their algorithms, TESA and TABU, based on one run per instance. To provide a reliable estimate of performance, we repeated our experiments 20 times with different random seeds, and report the average solutions on each instance. The best results are also indicated to establish bounds for future research. We adopted the same time limits as \cite{Sahin13}: 25.50 minutes per run for each instance with 75 pairs, 56.20 minutes for each instance with 100 pairs, and 95.90 minutes for each instance with 125 pairs. Tables \ref{nowak75}, \ref{nowak100} and \ref{nowak125} display the results on these instances. For each instance, the result of the best method is highlighted in boldface.

\begin{table}[!htbp]
\footnotesize
\centering
\caption{Results for the PDPSL with 75 pairs -- Instances from \cite{Nowak08} \\ Time limit set to 25.5 minutes per run}
\begin{tabular}{|c|c|cc|cc|cc|cc|}
\hline
& & \multicolumn{2}{c|}{\textbf{TESA}}  & \multicolumn{2}{c|}{\textbf{TABU}}  & \multicolumn{4}{c|}{\textbf{ILS--PDSL}}   \\ \hline
\textbf{Instance} & \textbf{BKS} & \textbf{1-Run} & \textbf{Gap(\%)} & \textbf{1-Run} & \textbf{Gap(\%)} & \textbf{Avg-20} & \textbf{Gap(\%)} & \textbf{Best-20} & \textbf{Gap(\%)}  \\ \hline
\textbf{75\_1A} & 3796.32 & 3830.12 & 0.89 & 3894.34 & 2.58 & \textbf{3786.30} & \textbf{-0.26} & 3727.32 & -1.82 \\ 
\textbf{75\_1B} & 3808.16 & 3857.12 & 1.29 & 3842.88 & 0.91 & \textbf{3764.84} & \textbf{-1.14} & 3683.20 & -3.28 \\ 
\textbf{75\_1C} & 3790.03 & 3810.50 & 0.54 & 3790.03 & 0.00 & \textbf{3767.15} & \textbf{-0.60} & 3686.44 & -2.73 \\ 
\textbf{75\_1D} & 3799.32 & 3799.32 & 0.00 & 3862.23 & 1.66 & \textbf{3755.47} & \textbf{-1.15} & 3707.17 & -2.43 \\ 
\textbf{75\_1E} & 3788.37 & 3868.96 & 2.13 & 3820.87 & 0.86 & \textbf{3781.96} & \textbf{-0.17} & 3719.38 & -1.82 \\ 
\textbf{75\_2A} & 3161.69 & 3313.48 & 4.80 & 3177.98 & 0.52 & \textbf{3104.55} & \textbf{-1.81} & 3044.34 & -3.71 \\ 
\textbf{75\_2B} & 3169.92 & 3296.36 & 3.99 & 3179.00 & 0.29 & \textbf{3102.96} & \textbf{-2.11} & 3069.97 & -3.15 \\ 
\textbf{75\_2C} & 3121.97 & 3203.25 & 2.60 & 3121.97 & 0.00 & \textbf{3085.99} & \textbf{-1.15} & 3040.38 & -2.61 \\ 
\textbf{75\_2D} & 3117.69 & 3266.42 & 4.77 & 3117.69 & 0.00 & \textbf{3074.07} & \textbf{-1.40} & 3010.40 & -3.44 \\ 
\textbf{75\_2E} & 3148.70 & 3332.59 & 5.84 & 3168.66 & 0.63 & \textbf{3109.04} & \textbf{-1.26} & 3075.34 & -2.33 \\ 
\textbf{75\_3A} & 3897.12 & 4058.37 & 4.14 & 3910.04 & 0.33 & \textbf{3892.26} & \textbf{-0.12} & 3817.39 & -2.05 \\ 
\textbf{75\_3B} & 3868.75 & 4172.42 & 7.85 & 3868.75 & 0.00 & \textbf{3860.70} & \textbf{-0.21} & 3771.77 & -2.51 \\ 
\textbf{75\_3C} & 3858.71 & 4090.65 & 6.01 & 3900.38 & 1.08 & \textbf{3866.62} & \textbf{0.20} & 3787.46 & -1.85 \\ 
\textbf{75\_3D} & 3845.05 & 4110.39 & 6.90 & 3888.20 & 1.12 & \textbf{3850.45} & \textbf{0.14} & 3762.61 & -2.14 \\ 
\textbf{75\_3E} & 3893.36 & 4052.23 & 4.08 & 3908.01 & 0.38 & \textbf{3826.36} & \textbf{-1.72} & 3733.67 & -4.10 \\ \hline
\textbf{Avg} & \textbf{} &  & 3.72 &  & 0.69 &  & \textbf{-0.85} & \textbf{} & -2.66 \\ \hline
 & & \multicolumn{2}{c|}{{Xeon}}  & \multicolumn{2}{c|}{{{Intel Core 2 Quad}}}  & \multicolumn{4}{c|}{{{Intel Core 2 Quad}}}  \\ 
\textbf{CPU} & & \multicolumn{2}{c|}{{2.4 GHz}}  & \multicolumn{2}{c|}{{2.4 GHz}}  & \multicolumn{4}{c|}{{2.4 GHz}}  \\ 
 & & \multicolumn{2}{c|}{{2 GB}}  & \multicolumn{2}{c|}{{4 GB}}  & \multicolumn{4}{c|}{{4 GB}}  \\ \hline
\end{tabular}
\label{nowak75}
\end{table}

\begin{table}[!htbp]
\footnotesize
\centering
\caption{Results for the PDPSL with 100 pairs -- Instances from \cite{Nowak08} \\ Time limit set to 56.2 minutes per run}
\begin{tabular}{|c|c|cc|cc|cc|cc|}
\hline
& & \multicolumn{2}{c|}{\textbf{TESA}}  & \multicolumn{2}{c|}{\textbf{TABU}}  & \multicolumn{4}{c|}{\textbf{ILS--PDSL}}   \\ \hline
\textbf{Instance} & \textbf{BKS} & \textbf{1-Run} & \textbf{Gap(\%)} & \textbf{1-Run} & \textbf{Gap(\%)} & \textbf{Avg-20} & \textbf{Gap(\%)} & \textbf{Best-20} & \textbf{Gap(\%)}  \\ \hline
\textbf{100\_1A} & 4992.59 & 5073.40 & 1.62 & 4992.59 & 0.00 & \textbf{4886.80} & \textbf{-2.12} & 4823.76 & -3.38 \\ 
\textbf{100\_1B} & 5036.55 & 5036.55 & 0.00 & 5042.30 & 0.11 & \textbf{4921.42} & \textbf{-2.29} & 4861.49 & -3.48 \\ 
\textbf{100\_1C} & 5015.09 & 5029.38 & 0.29 & 5015.09 & 0.00 & \textbf{4922.82} & \textbf{-1.84} & 4813.72 & -4.02 \\ 
\textbf{100\_1D} & 4996.08 & 5012.97 & 0.34 & 4996.08 & 0.00 & \textbf{4922.44} & \textbf{-1.47} & 4831.00 & -3.30 \\ 
\textbf{100\_1E} & 5015.26 & 5130.15 & 2.29 & 5015.26 & 0.00 & \textbf{4896.29} & \textbf{-2.37} & 4792.94 & -4.43 \\ 
\textbf{100\_2A} & 4204.28 & 4450.06 & 5.85 & 4258.49 & 1.29 & \textbf{4169.72} & \textbf{-0.82} & 4096.25 & -2.57 \\ 
\textbf{100\_2B} & 4306.73 & 4484.47 & 4.13 & 4306.73 & 0.00 & \textbf{4225.95} & \textbf{-1.88} & 4156.74 & -3.48 \\ 
\textbf{100\_2C} & 4215.07 & 4473.39 & 6.13 & 4259.09 & 1.04 & \textbf{4201.15} & \textbf{-0.33} & 4134.60 & -1.91 \\ 
\textbf{100\_2D} & 4244.77 & 4424.57 & 4.24 & 4267.37 & 0.53 & \textbf{4194.76} & \textbf{-1.18} & 4089.79 & -3.65 \\ 
\textbf{100\_2E} & 4228.82 & 4559.26 & 7.81 & 4228.82 & 0.00 & \textbf{4200.25} & \textbf{-0.68} & 4132.97 & -2.27 \\ 
\textbf{100\_3A} & 5126.71 & 5294.37 & 3.27 & 5126.71 & 0.00 & \textbf{4987.04} & \textbf{-2.72} & 4934.62 & -3.75 \\ 
\textbf{100\_3B} & 5084.70 & 5371.74 & 5.65 & 5161.29 & 1.51 & \textbf{5042.60} & \textbf{-0.83} & 4974.61 & -2.17 \\ 
\textbf{100\_3C} & 5075.45 & 5216.80 & 2.78 & 5098.71 & 0.46 & \textbf{5004.95} & \textbf{-1.39} & 4938.02 & -2.71 \\ 
\textbf{100\_3D} & 5106.32 & 5467.79 & 7.08 & 5106.32 & 0.00 & \textbf{5010.16} & \textbf{-1.88} & 4941.18 & -3.23 \\ 
\textbf{100\_3E} & 5076.14 & 5572.47 & 9.78 & 5076.14 & 0.00 & \textbf{5029.86} & \textbf{-0.91} & 4884.24 & -3.78 \\ \hline
\textbf{Avg} & \textbf{} &  & 4.08 &  & 0.33 & \textbf{} & \textbf{-1.51} & \textbf{} & -3.21 \\ \hline
 & & \multicolumn{2}{c|}{{Xeon}}  & \multicolumn{2}{c|}{{{Intel Core 2 Quad}}}  & \multicolumn{4}{c|}{{{Intel Core 2 Quad}}}  \\ 
\textbf{CPU} & & \multicolumn{2}{c|}{{2.4 GHz}}  & \multicolumn{2}{c|}{{2.4 GHz}}  & \multicolumn{4}{c|}{{2.4 GHz}}  \\ 
 & & \multicolumn{2}{c|}{{2 GB}}  & \multicolumn{2}{c|}{{4 GB}}  & \multicolumn{4}{c|}{{4 GB}}  \\ \hline
\end{tabular}
\label{nowak100}
\end{table}

\begin{table}[!htbp]
\footnotesize
\centering
\caption{Results for the PDPSL with 125 pairs -- Instances from \cite{Nowak08} \\ Time limit set to 95.9 minutes per run}
\begin{tabular}{|c|c|cc|cc|cc|cc|}
\hline
& & \multicolumn{2}{c|}{\textbf{TESA}}  & \multicolumn{2}{c|}{\textbf{TABU}}  & \multicolumn{4}{c|}{\textbf{ILS--PDSL}}   \\ \hline
\textbf{Instance} & \textbf{BKS} & \textbf{1-Run} & \textbf{Gap(\%)} & \textbf{1-Run} & \textbf{Gap(\%)} & \textbf{Avg-20} & \textbf{Gap(\%)} & \textbf{Best-20} & \textbf{Gap(\%)}  \\ \hline
\textbf{125\_1A} & 5950.44 & 6020.05 & 1.17 & 6002.15 & 0.87 & \textbf{5762.68} & \textbf{-3.16} & 5682.79 & -4.50 \\ 
\textbf{125\_1B} & 5938.94 & 5938.94 & 0.00 & 5998.06 & 1.00 & \textbf{5785.19} & \textbf{-2.59} & 5678.06 & -4.39 \\ 
\textbf{125\_1C} & 5933.69 & 5977.69 & 0.74 & 5933.69 & 0.00 & \textbf{5758.32} & \textbf{-2.96} & 5625.24 & -5.20 \\ 
\textbf{125\_1D} & 6060.85 & 6138.94 & 1.29 & 6083.59 & 0.38 & \textbf{5802.33} & \textbf{-4.27} & 5701.05 & -5.94 \\ 
\textbf{125\_1E} & 5906.34 & 6024.26 & 2.00 & 5906.34 & 0.00 & \textbf{5755.05} & \textbf{-2.56} & 5660.45 & -4.16 \\ 
\textbf{125\_2A} & 5396.85 & 5717.54 & 5.94 & 5444.23 & 0.88 & \textbf{5262.65} & \textbf{-2.49} & 5183.40 & -3.96 \\ 
\textbf{125\_2B} & 5456.91 & 5745.38 & 5.29 & 5460.81 & 0.07 & \textbf{5313.86} & \textbf{-2.62} & 5209.02 & -4.54 \\ 
\textbf{125\_2C} & 5412.81 & 5667.26 & 4.70 & 5412.81 & 0.00 & \textbf{5289.23} & \textbf{-2.28} & 5145.39 & -4.94 \\ 
\textbf{125\_2D} & 5475.40 & 5778.58 & 5.54 & 5494.71 & 0.35 & \textbf{5321.00} & \textbf{-2.82} & 5234.01 & -4.41 \\ 
\textbf{125\_2E} & 5419.02 & 5780.01 & 6.66 & 5419.02 & 0.00 & \textbf{5281.44} & \textbf{-2.54} & 5191.63 & -4.20 \\ 
\textbf{125\_3A} & 6237.20 & 6934.05 & 11.17 & 6252.24 & 0.24 & \textbf{6128.28} & \textbf{-1.75} & 6050.78 & -2.99 \\ 
\textbf{125\_3B} & 6300.04 & 6918.16 & 9.81 & 6300.04 & 0.00 & \textbf{6152.84} & \textbf{-2.34} & 6057.74 & -3.85 \\ 
\textbf{125\_3C} & 6324.66 & 6607.30 & 4.47 & 6332.93 & 0.13 & \textbf{6129.45} & \textbf{-3.09} & 6024.87 & -4.74 \\ 
\textbf{125\_3D} & 6317.05 & 7239.79 & 14.61 & 6359.16 & 0.67 & \textbf{6166.94} & \textbf{-2.38} & 6040.13 & -4.38 \\ 
\textbf{125\_3E} & 6257.16 & 6776.37 & 8.30 & 6277.38 & 0.32 & \textbf{6137.54} & \textbf{-1.91} & 6057.75 & -3.19 \\ \hline
\textbf{Avg} & \textbf{} &  & 5.45 &  & 0.33 &  & \textbf{-2.65} & \textbf{} & -4.36 \\ \hline
 & & \multicolumn{2}{c|}{{Xeon}}  & \multicolumn{2}{c|}{{{Intel Core 2 Quad}}}  & \multicolumn{4}{c|}{{{Intel Core 2 Quad}}}  \\ 
\textbf{CPU} & & \multicolumn{2}{c|}{{2.4 GHz}}  & \multicolumn{2}{c|}{{2.4 GHz}}  & \multicolumn{4}{c|}{{2.4 GHz}}  \\ 
 & & \multicolumn{2}{c|}{{2 GB}}  & \multicolumn{2}{c|}{{4 GB}}  & \multicolumn{4}{c|}{{4 GB}}  \\ \hline
\end{tabular}
\label{nowak125}
\end{table}

From these experiments, ILS--PDSL appears to produce solutions of higher quality than the TESA and TABU algorithms, as it was able to find better average results on all 45 instances. The largest improvements occur on the largest data sets.
Considering the average gaps, we observe negative values for every instance set ($-0.85\%$, $-1.51\%$ and~$-2.65\%$), meaning that the average solution quality of ILS--PDSL is better than the BKS in the literature. Finally, considering the best results out of 20 runs, we observe large improvements of the BKS ($3.41\%$ overall), with new best solutions for all 45 instances.

We conducted a Friedman test comparing the solution values for each instance to validate the statistical significance of the results. This test led to a value $p < 2.2 \times 10^{-16}$, which indicates a significant difference of performance. We also performed pairwise Wilcoxon tests to locate these differences which, as reported in Table \ref{wilcox}, support the existence of significant differences between all three methods: ILS--PDSL is significantly better than TABU, which is in turn significantly better than TESA.
 
\begin{table}[!htbp]
\renewcommand{\arraystretch}{1.2}
\footnotesize
\centering
\caption{Results of pairwise Wilcoxon tests\label{wilcox} -- Instances from \cite{Nowak08}}
\begin{tabular}{|c|c|}
\hline
\textbf{Algorithms}  &  \textbf{p-value} \\ \hline
\textbf{TESA and TABU}  &  $2.12 \times 10^{-10}$ \\
\textbf{ILS--PDSL and TESA} & $5.68 \times 10^{-14}$	 \\
\textbf{ILS--PDSL and TABU} & $5.68 \times 10^{-14}$ \\ \hline
\end{tabular}
\end{table}

\paragraph{Instances from \cite{Sahin13}.}

\cite{Sahin13} introduced a second set of instances and presented, for each instance with 50, 100 or 250 p-d pairs, the best solutions obtained by TABU over 20 runs. For the instances with 500 pairs, the authors presented the best solutions over five runs. As indicated by the authors in a private communication, the associated time values correspond to the average time of one run. These values also depend on the specific instance. As such, we have defined for each group of instances a termination criterion~$T_\textsc{max}$ which is smaller or equal to the average CPU time of TABU:  5~seconds for the instances with~50 service pairs, 40~seconds for the instances with 100 pairs, 5~minutes for the instances with 250 pairs, and 1~hour for the instances with 500 pairs. Tables \ref{ropke50}--\ref{ropke500} display the results of these experiments. In these tables, the solution quality of the best method is highlighted in boldface.

\begin{table}[!htbp]
\footnotesize
\centering
\caption{Results for the PDPSL with 50 pairs -- Instances from \cite{Sahin13} \\ Time limit set to 5 seconds per run}
\begin{tabular}{|c|cc|cc|cc|}
\hline
& \multicolumn{2}{c|}{\textbf{TABU}}  & \multicolumn{4}{c|}{\textbf{ILS--PDSL}}  \\ \hline
\textbf{Instance} & \textbf{T(s)} & \textbf{Best-20} & \textbf{Best-20} & \textbf{Gap(\%)} & \textbf{Avg-20} & \textbf{Gap(\%)} \\ \hline
\textbf{50A} & 4.5 & 16791.20 & \textbf{15481.36} & \textbf{-7.80} & 15913.01 & -5.23 \\ 
\textbf{50B} & 4.3 & 17115.50 & \textbf{15422.03} & \textbf{-9.89} & 15814.13 & -7.60 \\ 
\textbf{50C} & 4.5 & 14956.00 & \textbf{14131.43} & \textbf{-5.51} & 14591.86 & -2.43 \\ 
\textbf{50D} & 4.3 & 16290.00 & \textbf{14947.06} & \textbf{-8.24} & 15345.82 & -5.80 \\ 
\textbf{50E} & 7.6 & 11397.50 & \textbf{9517.49} & \textbf{-16.49} & 9895.06 & -13.18 \\ 
\textbf{50F} & 7.4 & 9532.59 & \textbf{8429.16} & \textbf{-11.58} & 8927.89 & -6.34 \\ 
\textbf{50G} & 6.2 & 9665.06 & \textbf{8820.07} & \textbf{-8.74} & 9175.39 & -5.07 \\ 
\textbf{50H} & 11.2 & 9199.58 & \textbf{7608.63} & \textbf{-17.29} & 7930.69 & -13.79 \\ 
\textbf{50I} & 5.8 & 14469.40 & \textbf{12864.70} & \textbf{-11.09} & 13235.73 & -8.53 \\ 
\textbf{50J} & 6.5 & 13200.20 & \textbf{11891.39} & \textbf{-9.92} & 12131.98 & -8.09 \\ 
\textbf{50K} & 2.3 & 12759.30 & \textbf{12337.42} & \textbf{-3.31} & 12594.40 & -1.29 \\ 
\textbf{50L} & 4.4 & 14867.80 & \textbf{13426.21} & \textbf{-9.70} & 13973.27 & -6.02 \\ \hline
\textbf{Avg} & 5.7 & \multicolumn{1}{l|}{} & \multicolumn{1}{l}{\textbf{}} & \textbf{-9.96} &  & -6.95 \\ \hline
\end{tabular}
\label{ropke50}
\end{table}

\begin{table}[!htbp]
\footnotesize
\centering
\caption{Results for the PDPSL with 100 pairs -- Instances from \cite{Sahin13} \\ Time limit set to 40 seconds per run}
\begin{tabular}{|c|cc|cc|cc|}
\hline
& \multicolumn{2}{c|}{\textbf{TABU}}  & \multicolumn{4}{c|}{\textbf{ILS--PDSL}}  \\ \hline
\textbf{Instance} & \textbf{T(s)} & \textbf{Best-20} & \textbf{Best-20} & \textbf{Gap(\%)} & \textbf{Avg-20} & \textbf{Gap(\%)} \\ \hline
\textbf{100A} & 25.1 & 27301.2 & \textbf{25398.19} & \textbf{-6.97} & 26268.75 & -3.78 \\ 
\textbf{100B} & 19.4 & 27090.1 & \textbf{25027.88} & \textbf{-7.61} & 26020.76 & -3.95 \\ 
\textbf{100C} & 34.0 & 27221.3 & \textbf{25319.76} & \textbf{-6.99} & 25833.95 & -5.10 \\ 
\textbf{100D} & 19.0 & 28574.7 & \textbf{26110.15} & \textbf{-8.62} & 27177.34 & -4.89 \\ 
\textbf{100E} & 74.7 & 15320 & \textbf{13498.17} & \textbf{-11.89} & 14022.84 & -8.47 \\ 
\textbf{100F} & 95.8 & 17574.2 & \textbf{13548.03} & \textbf{-22.91} & 13919.54 & -20.80 \\ 
\textbf{100G} & 50.1 & 14888.4 & \textbf{14508.21} & \textbf{-2.55} & 15062.04 & 1.17 \\ 
\textbf{100H} & 57.9 & 16259.7 & \textbf{14445.99} & \textbf{-11.15} & 15021.09 & -7.62 \\ 
\textbf{100I} & 32.4 & 24994.4 & \textbf{22603.21} & \textbf{-9.57} & 23292.98 & -6.81 \\ 
\textbf{100J} & 37.5 & 23025.5 & \textbf{21284.65} & \textbf{-7.56} & 21843.80 & -5.13 \\ 
\textbf{100K} & 30.4 & 24509 & \textbf{22435.89} & \textbf{-8.46} & 23248.32 & -5.14 \\ 
\textbf{100L} & 49.3 & 23994.7 & \textbf{20705.86} & \textbf{-13.71} & 21400.98 & -10.81 \\ \hline
\textbf{Avg} & 43.8 & \multicolumn{1}{l|}{} & \multicolumn{1}{l}{\textbf{}} & \textbf{-9.83} &  & -6.78 \\ \hline
\end{tabular}
\label{ropke100}
\end{table}

\begin{table}[!htbp]
\footnotesize
\centering
\caption{Results for the PDPSL with 250 pairs -- Instances from \cite{Sahin13} \\ Time limit set to 5 minutes per run}
\begin{tabular}{|c|cc|cc|cc|}
\hline
& \multicolumn{2}{c|}{\textbf{TABU}}  & \multicolumn{4}{c|}{\textbf{ILS--PDSL}}  \\ \hline
\textbf{Instance} & \textbf{T(s)} & \textbf{Best-20} & \textbf{Best-20} & \textbf{Gap(\%)} & \textbf{Avg-20} & \textbf{Gap(\%)} \\ \hline
\textbf{250A} & 287.3 & 58847.6 & \textbf{56857.45} & \textbf{-3.38} & 58821.37 & -0.04 \\ 
\textbf{250B} & 253.0 & 57559.1 & \textbf{55871.66} & \textbf{-2.93} & 57637.00 & 0.14 \\ 
\textbf{250C} & 299.5 & 57495.9 & \textbf{56483.36} & \textbf{-1.76} & 58107.41 & 1.06 \\ 
\textbf{250D} & 356.1 & 59396.7 & \textbf{57368.20} & \textbf{-3.42} & 59438.78 & 0.07 \\ 
\textbf{250E} & 3174.6 & 31736.8 & \textbf{28327.20} & \textbf{-10.74} & 29454.09 & -7.19 \\ 
\textbf{250F} & 1123.1 & 27596 & \textbf{24820.19} & \textbf{-10.06} & 25562.37 & -7.37 \\ 
\textbf{250G} & 1089.3 & 29421.8 & \textbf{26552.49} & \textbf{-9.75} & 27549.93 & -6.36 \\ 
\textbf{250H} & 939.5 & 31911.5 & \textbf{27326.55} & \textbf{-14.37} & 28656.36 & -10.20 \\ 
\textbf{250I} & 468.2 & 50154.8 & \textbf{48124.77} & \textbf{-4.05} & 50165.75 & 0.02 \\ 
\textbf{250J} & 448.8 & 53636.2 & \textbf{51119.88} & \textbf{-4.69} & 52868.55 & -1.43 \\ 
\textbf{250K} & 537.5 & 50084.4 & \textbf{46946.17} & \textbf{-6.27} & 49128.23 & -1.91 \\ 
\textbf{250L} & 392.3 & 54393.4 & \textbf{52580.00} & \textbf{-3.33} & 55067.03 & 1.24 \\ \hline
\textbf{Avg} & 780.8 & \multicolumn{1}{l|}{} & \multicolumn{1}{l}{\textbf{}} & \textbf{-6.23} &  & -2.66 \\ \hline
 \end{tabular}
\label{ropke250}
\end{table}

\begin{table}[!htbp]
\footnotesize
\centering
\caption{Results for the PDPSL with 500 pairs -- Instances from \cite{Sahin13} \\ Time limit set to 1 hour per run}
\begin{tabular}{|c|cc|cc|cc|}
\hline
& \multicolumn{2}{c|}{\textbf{TABU}}  & \multicolumn{4}{c|}{\textbf{ILS--PDSL}}  \\ \hline
\textbf{Instance} & \textbf{T(s)} & \textbf{Best-5} & \textbf{Best-5} & \textbf{Gap(\%)} & \textbf{Avg-5} & \textbf{Gap(\%)} \\ \hline
\textbf{500A} & 2124.6 & 106674 & \textbf{105536.28} & \textbf{-1.07} & 107176.11 & 0.47 \\ 
\textbf{500B} & 2374.2 & 110881 & \textbf{107657.18} & \textbf{-2.91} & 109636.75 & -1.12 \\ 
\textbf{500C} & 1985.8 & 109181 & \textbf{107676.77} & \textbf{-1.38} & 109991.16 & 0.74 \\ 
\textbf{500D} & 2247.0 & 109746 & \textbf{104432.98} & \textbf{-4.84} & 107220.16 & -2.30 \\ 
\textbf{500E} & 10860.4 & 63068.4 & \textbf{62322.13} & \textbf{-1.18} & 63411.06 & 0.54 \\ 
\textbf{500F} & 10815.8 & 68829.7 & \textbf{62951.49} & \textbf{-8.54} & 64701.06 & -6.00 \\ 
\textbf{500G} & 11101.8 & 70038.8 & \textbf{67147.93} & \textbf{-4.13} & 69658.51 & -0.54 \\ 
\textbf{500H} & 16763.8 & 60568.5 & \textbf{60489.14} & \textbf{-0.13} & 62636.44 & 3.41 \\ 
\textbf{500I} & 5075.4 & 93178.2 & 94264.57 & 1.17 & 97404.34 & 4.54 \\ 
\textbf{500J} & 4698.0 & 96984.8 & \textbf{94512.62} & \textbf{-2.55} & 97141.73 & 0.16 \\ 
\textbf{500K} & 4539.6 & 97429.5 & \textbf{96717.63} & \textbf{-0.73} & 98134.65 & 0.72 \\ 
\textbf{500L} & 5996.2 & 98102.7 & \textbf{95634.88} & \textbf{-2.52} & 97539.59 & -0.57 \\ \hline
\textbf{Avg} & 6548.6 & \multicolumn{1}{l|}{} & \multicolumn{1}{l}{\textbf{}} & \textbf{-2.40} &  & 0.00 \\ \hline
  \end{tabular}
\label{ropke500}
\end{table}

Since the best solution quality of TABU has been measured over multiple runs (20 or 5), the comparison is established with the best solution of ILS--PDSL over the same number of runs. When analyzing the tables, we observe that ILS--PDSL produces best solutions of higher quality than TABU (better than the BKS) on 47 instances out of 48. The significance of these improvements is again confirmed by a pairwise Wilcoxon test with a value $p = 2.35 \times 10^{-13}$. The magnitude of these improvements is also larger than on previous instances, with an improvement of $7.11\%$ on average (comparing best solutions together), which seems to indicate that these instances with a wider diversity of possible pickup and delivery locations are more difficult to solve, and remain challenging for future works.

\subsection{Metaheuristic -- Sensitivity analysis}

In order to examine the relative role of each component in the proposed heuristic, we started from the standard version of the algorithm and generated some alternative configurations by removing, in turn, a different neighborhood:
\begin{description}[nosep]
 \item[Base --] The standard configuration, with all local-search neighborhoods and the \emph{RCSP insertion};
 \item[\boldmath{$WN_1$} --] Base configuration without the \emph{PairSwap} neighborhood;
 \item[\boldmath{$WN_2$} --] Base configuration without the \emph{PairShift} neighborhood;
 \item[\boldmath{$WN_{34}$} --] Base configuration without the \emph{PickShift} and \emph{DelShift} neighborhoods;
 \item[\boldmath{$WN_5$} --] Base configuration without the \emph{BlockSwap} neighborhood;
 \item[\boldmath{$WN_6$} --] Base configuration without the \emph{BlockShift} neighborhood;
 \item[\boldmath{$WR$} --] Base configuration without the \emph{RCSP insertion} neighborhood. We note that the removal of the \emph{RCSP insertion} neighborhood forces the algorithm to work on a classic pickup and delivery problem, without possible split moves.
\end{description}

The resulting algorithms have been all tested on the instances of \cite{Nowak08}, performing five runs for each of the 45 data sets, and using the same termination criterion as in Section \ref{test11}. Table \ref{sensitivity_ILS--PDSL} displays, for each variant of the algorithm, the average gap for each set of instances (Gap-75, Gap-100 and Gap-125) as well as the average gap overall (Avg).

\begin{table}[htbp]
\footnotesize
\centering
\renewcommand{\arraystretch}{1.2}
\caption{Results for each configuration of the ILS--PDSL -- Instances from \cite{Nowak08}}
\begin{tabular}{|c|c|c|c|c|}
\hline
\textbf{Configuration} & \textbf{Gap-75(\%)} & \textbf{Gap-100(\%)} & \textbf{Gap-125(\%)} & \textbf{Avg(\%)} \\ \hline
\textbf{Base} & \textbf{-1.07} & \textbf{-1.82} & -2.57 & \textbf{-1.82} \\ 
\boldmath{$WN_1$} & -0.25 & -0.93 & -1.37 & -0.85 \\ 
\boldmath{$WN_2$}  & -0.98 & -1.58 & -2.52 & -1.70 \\ 
\boldmath{$WN_{34}$} & -0.75 & -1.55 & -2.46 & -1.59 \\ 
\boldmath{$WN_5$} & -0.98 & -1.74 & \textbf{-2.67} & -1.80 \\ 
\boldmath{$WN_6$} & -0.80 & -1.45 & -1.95 & -1.40 \\ 
\boldmath{$WR$} & 48.84 & 48.93 & 47.18 & 48.32 \\ \hline
\end{tabular}
\label{sensitivity_ILS--PDSL}
\end{table}

In this table, we observe that the Base configuration leads to the best overall gap ($-1.82\%$), as well as the best average gaps on the 75-pairs and 100-pairs instances. Still, the best average gap on the 125-pairs instances is attributed to the $WN_5$ variant, without the \emph{BlockSwap} neighborhood. This effect is possibly due to the variance of the solution quality of the algorithm on this relatively small sample of 15 instances, but it also demonstrates that some neighborhoods have a much larger impact than others. In decreasing order of importance, the most important neighborhood is the proposed \emph{RCSP insertion}, followed by the \emph{PairSwap} neighborhood, the \emph{BlockShift}, \emph{PickShift} and \emph{DelShift} neighborhoods, and then the others. The \emph{RCSP insertion}, in our context, is essential since it manages the optimization of the split loads.

\begin{figure}[hbtp]
\begin{center}
	\includegraphics[scale=0.52]{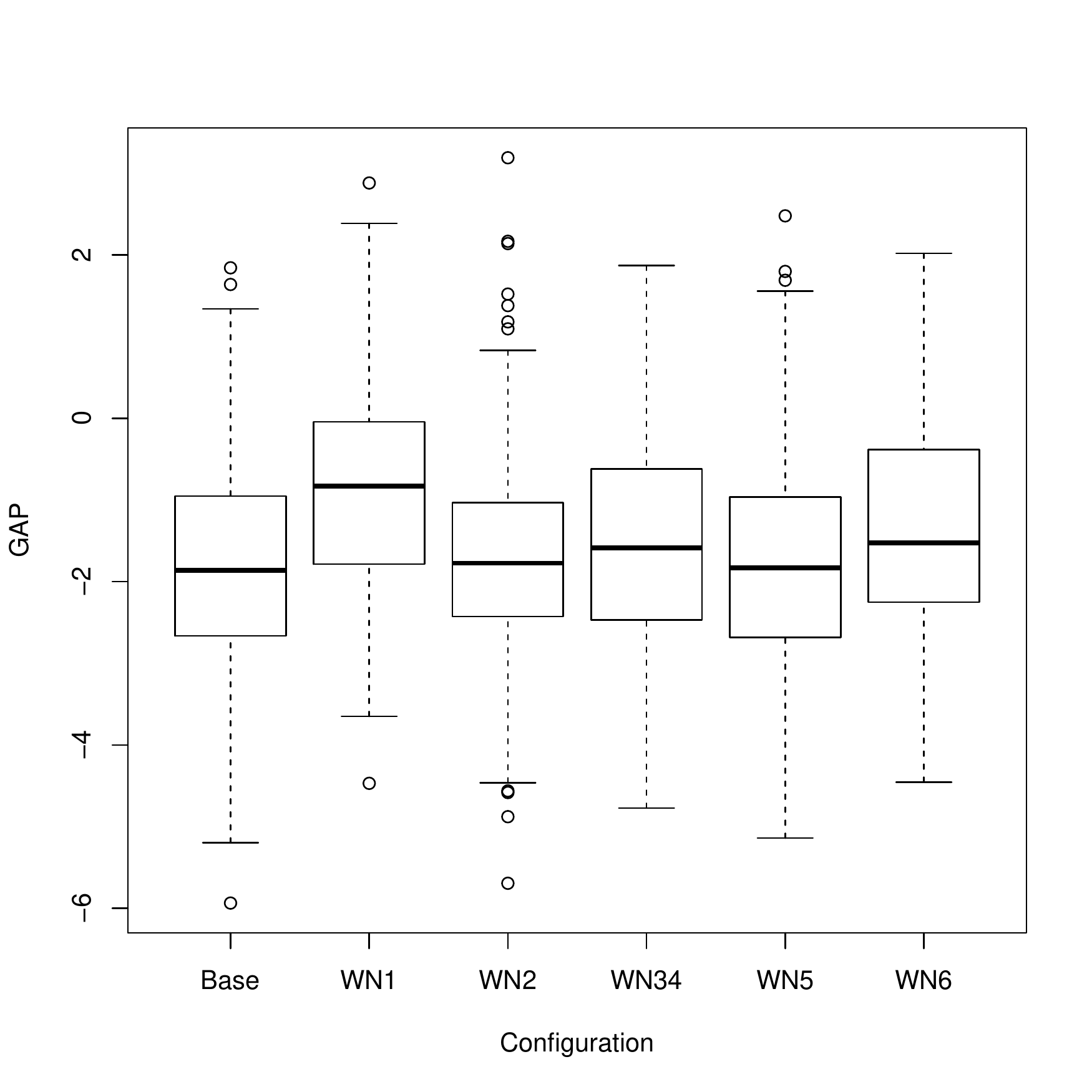}
	\caption{Box plot showing the gaps for each configuration of the ILS--PDSL\label{boxComp}}
\end{center}
\end{figure}

The gaps obtained by all ILS--PDSL variants (on all runs) can also be better observed by means of box plots, as in Figure \ref{boxComp}. In these box plots, represented without the results of $WR$ so as to enhance readability, we can observe the general superiority of the Base configuration. The removal of \emph{PairSwap} ({$WN_1$}) has a large negative impact on the final solutions, followed by the removal of \emph{BlockShift} ({$WN_6$}), the removal of \emph{PickShift} and \emph{DelShift} ({$WN_{34}$}), the removal of \emph{PairShift} ({$WN_2$}) and the removal of \emph{BlockSwap} ({$WN_5$}).

We performed a Friedman test based on the gap values of each algorithm to validate the previous observations. The test led to a value $p < 2.2 \times 10^{-16}$, demonstrating significant statistical differences. Then, we performed paired-sample Wilcoxon tests to compare the Base algorithm with all other algorithms. The results of these tests are reported in Table \ref{sensitivity_wilcox}.

\begin{table}[htbp]
\footnotesize
\centering
\caption{Results from paired-sample Wilcoxon tests with the Base algorithm\label{sensitivity_wilcox}}
\begin{tabular}{|c|c|}
\hline
\textbf{Algorithms}  &  \textbf{p-value} \\ \hline
\textbf{Base} -- \boldmath{$WN_1$}  &  $3.26 \times 10^{-16}$ \\
\textbf{Base} -- \boldmath{$WN_2$} & $0.18$ \\
\textbf{Base} -- \boldmath{$WN_{34}$} & $0.01$ \\
\textbf{Base} -- \boldmath{$WN_5$} & $0.78$ \\
\textbf{Base} -- \boldmath{$WN_6$} & $5.05 \times 10^{-05}$ \\ 
\textbf{Base} -- \boldmath{$WR$} & $< 2.2 \times 10^{-16}$ \\ \hline
\end{tabular}
\end{table}

These results confirm, with high confidence, the hypotheses that the Base algorithm produces results of significantly (better) quality than the $WN_1$, $WN_{34}$, $WN_6$ and $WR$ configurations, with p-values which are always smaller than a threshold of $0.05$. This highlights the importance of the neighborhoods which were deactivated in those configurations. A pairwise Wilcoxon test between the Base configuration and $WN_{2}$ and $WN_{5}$ led to p-values of $0.18$ and $0.78$, such that the significance of the difference of performance is not established in these cases. We can still reasonably conjecture that the associated neighborhoods (\emph{PairShift} and \emph{BlockSwap}) have a smaller impact, which would be better visible with additional runs and/or test instances. Besides, the CPU time consumption dedicated to these neighborhoods is very small, hence our choice to maintain them in the Base algorithm.

\subsection{Metaheuristic and exact -- Multiple vehicles and distance constraints}

As discussed in Sections \ref{sec:intro} and \ref{sec:literature}, the absence of distance constraints in the classical benchmark instances leads to the use of a single vehicle. To investigate real MPDPSL test cases, \myblue{we generated two sets of instances with distance constraints. The first set includes 40 small instances with 10 to 25 p-d pairs and a distance constraint $L = 300$. The second set} extends the 45 medium-size instances from \cite{Nowak08} with a distance constraint set to $L = 1000$. \myblue{All instances are available at \url{https://w1.cirrelt.ca/~vidalt/en/VRP-resources.html}.
The first set of instances allows to compare the results of the metaheuristic with some optimal solutions found by the B\&P. Due to their larger size, the instances of the second set are only solved heuristically.}

\myblue{Table \ref{tbl:exact-heur} presents the results obtained on the first set. The first group of columns report the average and best solutions found by the ILS-PDSL over 20 runs as well as the percentage gap between these two values. A time limit of one minute was imposed for each run. The second group of columns reports the results of the B\&P algorithm, considering a time limit of two hours. When this limit is reached, the time is reported as ``\textsc{tl}''. From left to right, the columns indicate the root node relaxation value, the time needed for the root node resolution, the final lower and upper bounds, the total CPU time, the percentage gap between the LB and UB, and finally the number of nodes explored in the branch-and-bound tree. For a few instances, indicated with ``--'', the B\&P could not complete the resolution of the root node within the time limit. When the exact method can prove optimality, the upper bound value is underlined. Finally, note that the B\&P receives the best solution of the ILS-PDSL as initial upper bound, therefore the columns UB is always smaller or equal to the best solution of the ILS-PDSL. In a few cases, the B\&P found a better UB during the execution. These solutions are highlighted in boldface.}

\myblue{These results show that the B\&P can solve instances of small and medium size. Out of the 40 instances of the first set, optimal solutions were found for 20 instances: for all instances with 10 p-d pairs, all but one with 15 pairs, and one with 20 pairs. For the last open instance with 15 pairs, the B\&P improved the upper bound found by the heuristic and attained a very small optimality gap ($0.4\%$). For the largest instances, the time needed for column generation becomes prohibitively high. In particular, the time limit of 2 hours was attained during the resolution of the root node on two instances with 25 p-d pairs. In the other cases, the B\&P still manages to find decent-quality lower bounds, at most 14.7\% away from the heuristic upper bound.

\myblue{
Considering the metaheuristic results, we observe that 18 out of the 20 known optimal solutions were found in at least one run. The average percentage deviation between the average and best solutions of ILS--PDSL, over the complete set of instances, amounts to 0.98\%, therefore illustrating the good stability of the method. Similarly, the deviation between the average solutions and the lower bound found by the B\&P method (eliminating instances 25-2 and 25-5) amounts to 4.65\%. This guarantees a good proximity between the solutions of ILS--PDSL and the true optima, even if this estimate is naturally pessimistic due to the gap between the LB and the optima.}
}

\begin{table}[!htbp]
\footnotesize
\centering
\caption{\myblue{Results for the MPDPSL -- Small instances \\ ILS-PDSL time limit set to 1 minute per run, B\&P time limit set to 2 hours}}
\myblue{
\begin{tabular}{|c|ccc|ccccccc|}\hline
\multirow{2}{*}{\textbf{Instance}} & \multicolumn{3}{c|}{\textbf{ILS-PDSL}} & \multicolumn{7}{c|}{\textbf{Branch-and-Price}} \\
 & \textbf{Best-20} & \textbf{Avg-20} & \textbf{Gap(\%)} & \textbf{LB$_0$} & \textbf{T$_{0}$(s)} & \textbf{LB} & \textbf{UB} & \textbf{T(s)} & \textbf{Gap(\%)} & \textbf{Nodes} \\\hline
\textbf{10-1} & 953 & 953.0 & 0.0 & 861.2 & 0.3 & 953.0 & \underline{953} & 7.3 & 0.0 & 23 \\
\textbf{10-2} & 1020 & 1021.8 & 0.2 & 907.7 & 2.5 & 1020.0 & \underline{1020} & 42.8 & 0.0 & 29 \\
\textbf{10-3} & 968 & 970.7 & 0.3 & 787.5 & 1.2 & 968.0 & \underline{968} & 1318.7 & 0.0 & 1865 \\
\textbf{10-4} & 979 & 979.0 & 0.0 & 851.4 & 1.0 & 979.0 & \underline{979} & 518.3 & 0.0 & 681 \\
\textbf{10-5} & 1026 & 1026.0 & 0.0 & 891.6 & 0.2 & 1026.0 & \underline{1026} & 41.7 & 0.0 & 293 \\
\textbf{10-6} & 742 & 742.0 & 0.0 & 653.7 & 0.5 & 742.0 & \underline{742} & 5.9 & 0.0 & 15 \\
\textbf{10-7} & 1019 & 1019.0 & 0.0 & 902.7 & 0.6 & 1019.0 & \underline{1019} & 15.8 & 0.0 & 25 \\
\textbf{10-8} & 818 & 818.0 & 0.0 & 725.9 & 0.6 & 818.0 & \underline{818} & 19.4 & 0.0 & 31 \\
\textbf{10-9} & 765 & 765.0 & 0.0 & 680.6 & 3.0 & 765.0 & \underline{765} & 161.6 & 0.0 & 89 \\
\textbf{10-10} & 1058 & 1058.0 & 0.0 & 1008.7 & 0.1 & 1058.0 & \underline{1058} & 3.7 & 0.0 & 29 \\
\textbf{15-1} & 1346 & 1379.2 & 2.5 & 1188.0 & 2.4 & 1346.0 & \underline{1346} & 165.6 & 0.0 & 65 \\
\textbf{15-2} & 1116 & 1116.0 & 0.0 & 1073.0 & 40.9 & 1116.0 & \underline{1116} & 666.8 & 0.0 & 29 \\
\textbf{15-3} & 1184 & 1194.6 & 0.9 & 1152.8 & 113.2 & 1175.5 & \textbf{1180} & \textsc{tl} & 0.4 & 122 \\
\textbf{15-4} & 1297 & 1302.6 & 0.4 & 1124.3 & 5.2 & 1297.0 & \underline{1297} & 3308.3 & 0.0 & 1029 \\
\textbf{15-5} & 1028 & 1028.0 & 0.0 & 898.7 & 13.3 & 1028.0 & \underline{1028} & 4533.0 & 0.0 & 639 \\
\textbf{15-6} & 1155 & 1157.8 & 0.2 & 1085.7 & 21.6 & 1140.0 & \textbf{\underline{1140}} & 249.9 & 0.0 & 13 \\
\textbf{15-7} & 1102 & 1116.9 & 1.4 & 1076.8 & 96.2 & 1102.0 & \underline{1102} & 857.9 & 0.0 & 13 \\
\textbf{15-8} & 1145 & 1145.3 & 0.0 & 1107.5 & 11.8 & 1145.0 & \underline{1145} & 322.4 & 0.0 & 43 \\
\textbf{15-9} & 1150 & 1154.4 & 0.4 & 1117.1 & 84.0 & 1150.0 & \underline{1150} & 3060.4 & 0.0 & 49 \\
\textbf{15-10} & 1060 & 1060.0 & 0.0 & 1010.1 & 27.5 & 1060.0 & \underline{1060} & 589.3 & 0.0 & 33 \\
\textbf{20-1} & 1350 & 1351.1 & 0.1 & 1216.8 & 478.5 & 1272.1 & 1350 & \textsc{tl} & 6.1 & 49 \\
\textbf{20-2} & 1454 & 1457.9 & 0.3 & 1333.9 & 69.8 & 1452.7 & 1454 & \textsc{tl} & 0.1 & 215 \\
\textbf{20-3} & 861 & 862.2 & 0.1 & 856.7 & 1210.9 & 861.0 & \underline{861} & 2999.7 & 0.0 & 3 \\
\textbf{20-4} & 1348 & 1350.8 & 0.2 & 1223.4 & 451.7 & 1267.4 & 1348 & \textsc{tl} & 6.4 & 37 \\
\textbf{20-5} & 1327 & 1337.0 & 0.7 & 1177.7 & 3117.5 & 1177.7 & 1327 & \textsc{tl} & 12.7 & 3 \\
\textbf{20-6} & 1614 & 1625.3 & 0.7 & 1447.3 & 127.0 & 1546.4 & 1614 & \textsc{tl} & 4.4 & 99 \\
\textbf{20-7} & 1399 & 1399.5 & 0.0 & 1297.7 & 2947.8 & 1297.7 & 1399 & \textsc{tl} & 7.8 & 3 \\
\textbf{20-8} & 1299 & 1329.3 & 2.3 & 1118.6 & 653.5 & 1143.3 & 1299 & \textsc{tl} & 13.6 & 13 \\
\textbf{20-9} & 1469 & 1588.0 & 8.1 & 1316.3 & 428.1 & 1369.4 & 1469 & \textsc{tl} & 7.3 & 17 \\
\textbf{20-10} & 1429 & 1460.5 & 2.2 & 1366.9 & 2379.3 & 1380.9 & 1429 & \textsc{tl} & 3.5 & 6 \\
\textbf{25-1} & 1887 & 1908.8 & 1.2 & 1619.3 & 2022.1 & 1665.7 & 1887 & \textsc{tl} & 13.3 & 6 \\
\textbf{25-2} & 1692 & 1748.4 & 3.3 & -- & -- & -- & -- & \textsc{tl} & -- & -- \\
\textbf{25-3} & 1549 & 1555.2 & 0.4 & 1391.0 & 2678.4 & 1391.0 & 1549 & \textsc{tl} & 11.4 & 3 \\
\textbf{25-4} & 1675 & 1685.5 & 0.6 & 1581.1 & 2621.1 & 1587.5 & 1675 & \textsc{tl} & 5.5 & 5 \\
\textbf{25-5} & 1415 & 1428.9 & 1.0 & -- & -- & -- & -- & \textsc{tl} & -- & -- \\
\textbf{25-6} & 1882 & 1923.3 & 2.2 & 1687.3 & 1425.6 & 1692.7 & 1882 & \textsc{tl} & 11.2 & 6 \\
\textbf{25-7} & 1487 & 1604.3 & 7.9 & 1442.5 & 4089.6 & 1442.5 & 1487 & \textsc{tl} & 3.1 & 3 \\
\textbf{25-8} & 1429 & 1446.6 & 1.2 & 1377.5 & 5308.6 & 1377.5 & 1429 & \textsc{tl} & 3.7 & 3 \\
\textbf{25-9} & 1613 & 1616.1 & 0.2 & 1400.9 & 1032.5 & 1422.1 & 1613 & \textsc{tl} & 13.4 & 8 \\
\textbf{25-10} & 1802 & 1802.6 & 0.0 & 1538.5 & 2335.8 & 1571.6 & 1802 & \textsc{tl} & 14.7 & 4 \\
\hline
\end{tabular}}
\label{tbl:exact-heur}
\end{table}

Tables~\mbox{\ref{mpdpsl75}--\ref{mpdpsl125}} display the results \myblue{of the metaheuristic for the second instance set, using the same time limits as in Section~\ref{test11}}: the average solution in the 20 executions (Avg-20), the average gap (Gap), the best solution in the 20 executions (Best-20), and the standard deviation related to the gaps (Std Dev) for each instance.

\begin{table}[!htbp]
\footnotesize
\centering
\caption{Results for the MPDPSL with 75 pairs -- Instances from \cite{Nowak08} \\ Time limit set to 25.50 mins per run}
\begin{tabular}{|c|c|cc|c|}
\hline
\textbf{Instance} & \textbf{Best-20} & \textbf{Avg-20} & \textbf{Gap(\%)} & \textbf{Std Dev(\%)} \\ \hline
\textbf{75\_1A} & 3782.93 & 3865.12 & 2.17 & 1.66 \\ 
\textbf{75\_1B} & 3747.57 & 3836.27 & 2.37 & 1.62 \\ 
\textbf{75\_1C} & 3793.62 & 3864.07 & 1.86 & 1.17 \\ 
\textbf{75\_1D} & 3765.05 & 3835.80 & 1.88 & 1.15 \\ 
\textbf{75\_1E} & 3757.36 & 3835.99 & 2.09 & 1.44 \\ 
\textbf{75\_2A} & 3097.22 & 3200.18 & 3.32 & 1.23 \\ 
\textbf{75\_2B} & 3123.89 & 3181.35 & 1.84 & 1.04 \\ 
\textbf{75\_2C} & 3110.82 & 3174.16 & 2.04 & 1.19 \\ 
\textbf{75\_2D} & 3097.16 & 3163.17 & 2.13 & 1.45 \\ 
\textbf{75\_2E} & 3118.38 & 3192.46 & 2.38 & 1.59 \\ 
\textbf{75\_3A} & 3866.08 & 3981.00 & 2.97 & 1.32 \\ 
\textbf{75\_3B} & 3855.72 & 3976.12 & 3.12 & 1.61 \\ 
\textbf{75\_3C} & 3886.66 & 3959.61 & 1.88 & 1.07 \\ 
\textbf{75\_3D} & 3870.89 & 3944.61 & 1.90 & 1.16 \\ 
\textbf{75\_3E} & 3828.10 & 3958.88 & 3.42 & 1.55 \\ \hline
\textbf{Avg} &  &  & 2.36 & 1.35 \\ \hline
\end{tabular}
\label{mpdpsl75}
\end{table}

\begin{table}[!htbp]
\footnotesize
\centering
\caption{Results for the MPDPSL with 100 pairs -- Instances from \cite{Nowak08} \\ Time limit set to 56.20 mins per run}
\begin{tabular}{|c|c|cc|c|}
\hline
\textbf{Instance} & \textbf{Best-20} & \textbf{Avg-20} & \textbf{Gap(\%)} & \textbf{Std Dev(\%)} \\ \hline
\textbf{100\_1A} & 4920.25 & 4993.39 & 1.49 & 1.02 \\ 
\textbf{100\_1B} & 4940.53 & 5029.61 & 1.80 & 1.08 \\ 
\textbf{100\_1C} & 4903.04 & 5010.18 & 2.19 & 1.17 \\ 
\textbf{100\_1D} & 4928.88 & 5012.28 & 1.69 & 0.86 \\ 
\textbf{100\_1E} & 4869.26 & 4977.90 & 2.23 & 1.05 \\ 
\textbf{100\_2A} & 4212.57 & 4270.83 & 1.38 & 0.88 \\ 
\textbf{100\_2B} & 4226.56 & 4304.39 & 1.84 & 1.13 \\ 
\textbf{100\_2C} & 4213.68 & 4281.58 & 1.61 & 1.06 \\ 
\textbf{100\_2D} & 4217.27 & 4305.97 & 2.10 & 1.27 \\ 
\textbf{100\_2E} & 4186.25 & 4275.55 & 2.13 & 1.16 \\ 
\textbf{100\_3A} & 4982.29 & 5131.18 & 2.99 & 1.31 \\ 
\textbf{100\_3B} & 5057.84 & 5189.52 & 2.60 & 1.38 \\ 
\textbf{100\_3C} & 5031.36 & 5137.47 & 2.11 & 1.38 \\ 
\textbf{100\_3D} & 5049.84 & 5152.46 & 2.03 & 1.35 \\ 
\textbf{100\_3E} & 5029.86 & 5144.86 & 2.29 & 1.46 \\ \hline
\textbf{Avg} &  &  & 2.03 & 1.17 \\ \hline
\end{tabular}
\label{mpdpsl100}
\end{table}

\begin{table}[!htbp]
\footnotesize
\centering
\caption{Results for the MPDPSL with 125 pairs -- Instances from \cite{Nowak08} \\ Time limit set to 95.90 mins per run}
\begin{tabular}{|c|c|cc|c|}
\hline
\textbf{Instance} & \textbf{Best-20} & \textbf{Avg-20} & \textbf{Gap(\%)} & \textbf{Std Dev(\%)} \\ \hline
\textbf{125\_1A} & 5794.79 & 5888.54 & 1.62 & 0.96 \\ 
\textbf{125\_1B} & 5880.96 & 5966.92 & 1.46 & 0.91 \\ 
\textbf{125\_1C} & 5738.87 & 5881.25 & 2.48 & 1.22 \\ 
\textbf{125\_1D} & 5738.32 & 5945.99 & 3.62 & 1.49 \\ 
\textbf{125\_1E} & 5822.20 & 5915.08 & 1.60 & 0.89 \\ 
\textbf{125\_2A} & 5310.49 & 5419.77 & 2.06 & 0.87 \\ 
\textbf{125\_2B} & 5361.25 & 5476.77 & 2.15 & 1.11 \\ 
\textbf{125\_2C} & 5357.90 & 5417.37 & 1.11 & 0.66 \\ 
\textbf{125\_2D} & 5331.69 & 5458.60 & 2.38 & 0.98 \\ 
\textbf{125\_2E} & 5339.33 & 5443.11 & 1.94 & 1.18 \\ 
\textbf{125\_3A} & 6177.11 & 6283.43 & 1.72 & 1.22 \\ 
\textbf{125\_3B} & 6205.87 & 6343.14 & 2.21 & 1.17 \\ 
\textbf{125\_3C} & 6230.02 & 6312.84 & 1.33 & 0.92 \\ 
\textbf{125\_3D} & 6181.96 & 6351.26 & 2.74 & 1.45 \\ 
\textbf{125\_3E} & 6128.42 & 6313.40 & 3.02 & 1.52 \\ \hline
\textbf{Avg} &  &  & 2.10 & 1.10 \\ \hline
\end{tabular}
\label{mpdpsl125}
\end{table}

These results aim to provide a useful base for comparisons with new algorithms in the future. They also reflect the difficulty of the problems, since small deviations related to the best solutions are usually good  indications of performance. From these experiments, we observe that the average gap remains moderate: $2.36\%$ for the 75-pairs set, $2.03\%$ for the 100-pairs set and $2.10\%$ for the 125-pairs set. These values are sensibly higher than those of the single-vehicle experiments, with average gaps of $1.86\%$, $1.75\%$ and $1.79\%$, respectively (when computed relatively to the best solutions of 20 runs). As such, the new MPDPSL instances appear to be more challenging, and would deserve further attention in the coming years.
Finally, the solutions of these larger multi-vehicle instances with distance constraints, and their single-vehicle counterpart, contain a high proportion of split loads (55.69\% and 58.52\%, respectively), likely due to the fact that each p-d pair occupies between 51\% to 60\% of the truck capacity.

\section{Conclusions}
\label{Conclusao}

In this article, we have considered the multi-vehicle one-to-one pickup and delivery problem with split loads (MPDPSL). Because this problem combines pickups and deliveries with split deliveries, solving it is a challenging task. \myblue{In particular, since the number of visits in a solution may grow exponentially with the instance size, no flow-based formulation with a polynomial number of variables can represent the problem. For local-search based heuristics,} the sequencing and split deliveries decision subsets are also very interdependent, such that various neighborhoods must be designed to jointly modify some of these decisions. 

To address these challenges, we proposed a \myblue{branch-and-price method as well as} a conceptually simple ILS, based on classic neighborhoods for pickup-and-delivery problems. Moreover, to efficiently manage the split deliveries, we introduced an exponential-sized neighborhood, which iteratively optimizes the pickup-and-delivery locations and splits for each service, and can be efficiently explored in pseudo-polynomial time.
The performance of the proposed \myblue{methods} has been validated through extensive computational experiments. For the classical single-vehicle problem instances, \myblue{our heuristic} outperforms all existing algorithms in similar computational time, and finds new best known solutions for 92 out of 93 instances. We also proposed new multi-vehicle problem instances and solutions for future comparisons. \myblue{For 20 instances, the branch-and-price algorithm could produce optimal solutions, and good-quality lower bounds were otherwise generated for the majority of small and medium instances.}

Overall, \myblue{our research on metaheuristics} takes place in a general research line which aims at progressing towards an intelligent search and exploration of larger neighborhoods via efficient dynamic-programming techniques, in contrast with the brute-force enumeration of simpler neighborhoods. The MPDPSL is a very challenging problem in this regard. For future research, we suggest to keep on generalizing these neighborhoods and their exploration techniques, as well as extending the methodology to a wider range of difficult vehicle routing variants. \myblue{For exact methods, many research avenues are also open. Similarly to the research conducted on the classical split delivery problem, new structural properties of MPDPSL optimal solutions and polyhedral results may be essential to trigger new methodological progress.}

\section*{Acknowledgements}

This research was partially supported by the CNPQ (grants 201554/2014-3, 308498/2015-1, 425962/2016-4 and 307915/2016-6), as well as CAPES and FAPEMIG, Brazil.

\bibliographystyle{elsarticle-harv} 
\bibliography{referencias}

\end{document}